\documentclass[10pt,twocolumn,letterpaper]{article}
\usepackage{bm}
\usepackage{mathrsfs}
\usepackage{multirow}
\usepackage{siunitx}
\newcolumntype{Y}{S[table-format=2.2, parse-numbers=false]@{\,/\,}S[table-format=1.2, parse-numbers=false]}
\usepackage[pagenumbers]{cvpr} 

\definecolor{cvprblue}{rgb}{0.21,0.49,0.74}
\usepackage[pagebackref,breaklinks,colorlinks,allcolors=cvprblue]{hyperref}

\def\paperID{19984} 
\def\confName{CVPR}
\def\confYear{2026}

\title{ShadowGS: Shadow-Aware 3D Gaussian Splatting for Satellite Imagery}

\author{Feng Luo,Hongbo Pan\textsuperscript{*},Xiang Yang,Baoyu Jiang,Fengqing Liu,Tao Huang\\
\\
Central South University\\
\\
{\tt\small 245007016@csu.edu.cn, \tt\small hongbopan@csu.edu.cn} 
}

\begin{document}
\maketitle

\renewcommand{\thefootnote}{*}
\footnotetext{Corresponding author.}
\begin{abstract}
3D Gaussian Splatting (3DGS) has emerged as a novel paradigm for 3D reconstruction from satellite imagery. However, in multi-temporal satellite images, prevalent shadows exhibit significant inconsistencies due to varying illumination conditions. To address this, we propose ShadowGS, a novel framework based on 3DGS. It leverages a physics-based rendering equation from remote sensing, combined with an efficient ray marching technique, to precisely model geometrically consistent shadows while maintaining efficient rendering. Additionally, it effectively disentangles different illumination components and apparent attributes in the scene. Furthermore, we introduce a shadow consistency constraint that significantly enhances the geometric accuracy of 3D reconstruction. We also incorporate a novel shadow map prior to improve performance with sparse-view inputs. Extensive experiments demonstrate that ShadowGS outperforms current state-of-the-art methods in shadow decoupling accuracy, 3D reconstruction precision, and novel view synthesis quality, with only a few minutes of training. ShadowGS exhibits robust performance across various settings, including RGB, pansharpened, and sparse-view satellite inputs.
\end{abstract}    
\section{Introduction}
\label{sec:intro}

High-resolution optical satellites capture large-scale, sub-meter imagery of the Earth's surface from orbital altitudes. Compared to close-range or UAV platforms, satellites provide extensive spatial coverage at lower acquisition costs, making them highly valuable for large-scale 3D reconstruction~\cite{ref14}, digital twins, and smart city applications~\cite{ref15}. With the ongoing launch of new-generation sub-meter satellites, imagery availability has increased significantly, further establishing satellite-based 3D reconstruction as an important research area in computer vision.

\begin{figure}[htb]
	\centering
	\includegraphics[width=\linewidth]{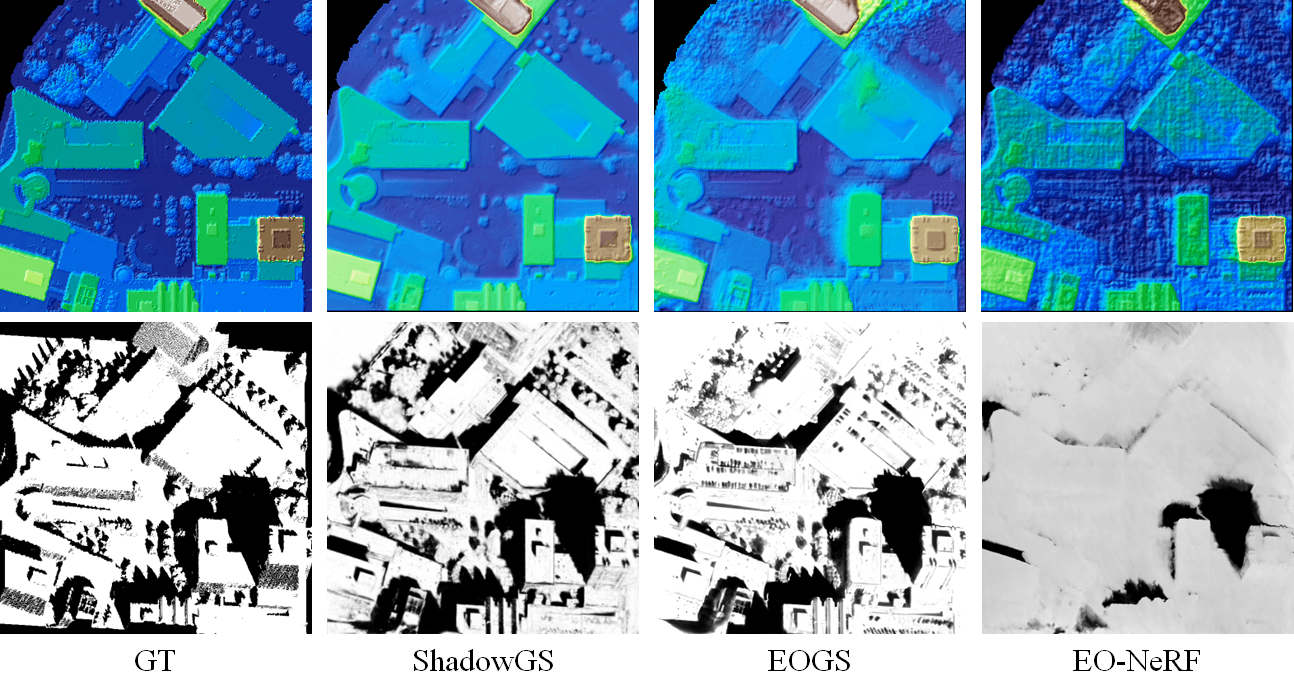}
	\caption{ShadowGS reconstructs 3D geometry with consistent shadow modeling from multi-temporal satellite imagery. The top row displays reconstructed DSMs while the bottom row shows shadow decomposition results. Compared to EO-NeRF~\cite{ref8} and EOGS~\cite{ref11}, our method produces superior reconstruction quality with sharper edges, richer details, smoother surfaces, and shadows that align precisely with scene geometry.}
	\label{fig1}
\end{figure}
However, due to orbital constraints, satellite images are often acquired at different times and from limited viewpoints, and frequently contain shadows caused by occluded sunlight. Although shadows can provide valuable 3D cues, their strong inconsistency across multi-temporal images poses significant challenges for reconstruction. Traditional multi-view stereo (MVS) methods~\cite{ref1,ref2,ref3,ref4} generally assume simultaneous image capture and struggle with strong appearance variations across time. Neural Radiance Fields (NeRF)~\cite{ref5} have shown promise for multi-temporal satellite imagery, yet limitations remain. For instance, S-NeRF~\cite{ref6} and SatNeRF~\cite{ref7} use MLPs to model shadow features related to sun position but fail to incorporate geometric context, leading to inaccurate shadow separation. EO-NeRF~\cite{ref8} models geometry-dependent shadows but lacks strict constraints between shadow and geometry, causing substantial shadow information to be entangled within geometric representations. Furthermore, NeRF-based methods typically suffer from high computational cost and slow inference.

3D Gaussian Splatting (3DGS)~\cite{ref9}, with its efficient rasterization-based rendering, has recently emerged as a promising alternative for satellite 3D reconstruction. However, rasterization is inherently local and struggles to model global effects like shadows. Existing 3DGS adaptations have attempted to handle multi-temporal shadow inconsistencies in different ways. SatGS~\cite{ref10} uses an MLP to estimate solar visibility per Gaussian but ignores geometric relationships, while EOGS~\cite{ref11} introduces shadow mapping to model geometry-aware shadows, though it remains an approximation and tends to produce aliasing artifacts.

To address these issues, we propose ShadowGS, a 3DGS-based framework that disentangles geometry (elevation and normals), appearance (albedo), and illumination (direct sunlight, skylight, and near-surface reflection) from multi-temporal satellite images, enabling rendering under arbitrary views and lighting. Specifically, ShadowGS assigns two sets of spherical harmonic (SH) coefficients to each Gaussian to represent albedo and near-surface reflection, while a global low-order SH models skylight. For geometrically consistent shadows, we cast rays from each Gaussian toward the sun and employ hardware-accelerated ray marching to determine occlusions and compute solar visibility. A physics-based rendering equation is then applied after rasterization to compose the final pixel color from albedo and multiple illumination components. To improve geometry, we derive Gaussian normals and depth contributions via ray-Gaussian intersection and enforce a depth-normal consistency constraint. We further introduce a shadow consistency constraint, which requires that when the camera view aligns with the sun direction, object shadows should be fully self-occluded---i.e., the rendered shadow map should be entirely lit. This encourages Gaussians to align closely with true surfaces and converge to higher opacity. In sparse-view settings, we integrate a pre-trained shadow detection network~\cite{ref16} to provide shadow map priors that guide optimization under limited inputs.

Experiments on the DFC2019~\cite{ref12, ref13} and IARPA~\cite{ref61} datasets show that ShadowGS accurately models geometry-aware shadows and outperforms existing methods in shadow disentanglement, 3D reconstruction accuracy, and novel view synthesis---all within minutes of training. Our main contributions are:
\begin{itemize}
    \item An efficient ray-marching-based shadow computation method for satellite imagery that models geometry-consistent shadows while maintaining high rendering efficiency.
    \item A remote-sensing physics-based rendering equation that effectively disentangles illumination and appearance.
    \item A shadow consistency constraint that significantly improves reconstruction quality.
    \item Integration of a shadow map prior to enhance performance under sparse-view conditions.
\end{itemize}


\section{Related Work}
\label{sec:Related}

\subsection{Shadow in Remote Sensing}

Shadows result from the occlusion of light propagation, revealing interactions among light sources, scene geometry, and object spatial relationships~\cite{ref17}. They are prevalent in remote sensing imagery and present dual characteristics: while often degrading image information and hampering visual interpretation, object classification, and quantitative inversion, they simultaneously offer valuable geometric cues. Prior work has leveraged shadows for building height estimation~\cite{ref22, ref23}, scene geometry recovery~\cite{ref18, ref19}, illumination direction estimation~\cite{ref20}, and camera calibration~\cite{ref21}.

Accurate shadow detection and removal are therefore crucial. Early methods primarily relied on handcrafted features and traditional machine learning~\cite{ref24, ref25}. Recent advances in deep learning have substantially improved performance in both detection and removal~\cite{ref16, ref26,ref27,ref28}. In remote sensing, the introduction of the AISD dataset~\cite{ref29} spurred the development of specialized detectors~\cite{ref30, ref31}. Notably, SEO~\cite{ref32} recently released a large-scale, high-resolution dataset containing multi-temporal and multi-view WorldView-3 imagery, along with geo-registered shadow masks and aligned LiDAR DSMs. Shadow detection networks trained on such data have been used to supervise EO-NeRF~\cite{ref8}, demonstrating the potential of shadow priors in enhancing radiance field methods.

\subsection{NeRF for Satellite Images}

NeRF\cite{ref5} model scenes as continuous volumetric representations using fully-connected networks. By mapping 3D coordinates and viewing directions to volume density and view-dependent color, and employing volume rendering, NeRF can optimize scene representations from images with known poses while handling complex appearance changes.

\begin{figure*}[ht]
	\centering
	\includegraphics[width=\linewidth]{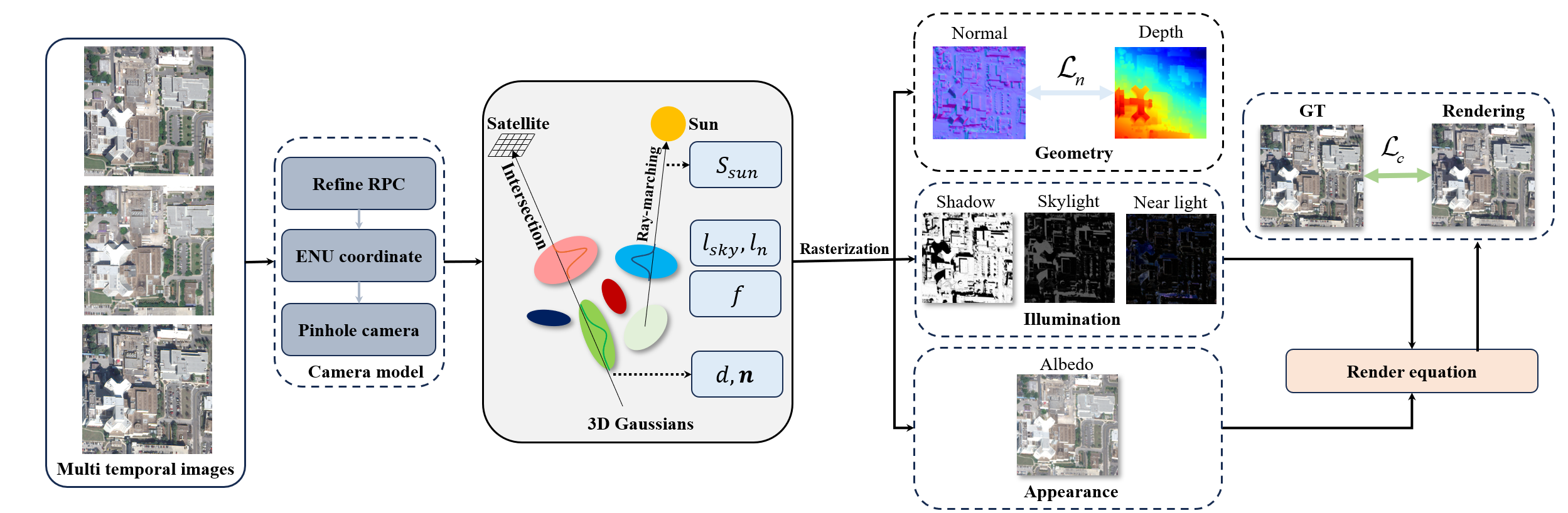}
	\caption{The overall pipeline of ShadowGS.}
	\label{fig2}
\end{figure*}

In remote sensing, NeRF has been adapted to address illumination inconsistencies, shadows, and transient objects in multi-temporal data. S-NeRF~\cite{ref6} first introduced NeRF to satellite photogrammetry, using a lighting model to decouple albedo and irradiance. Sat-NeRF~\cite{ref7} incorporated rational polynomial camera models and transient embeddings to handle dynamic elements. EO-NeRF~\cite{ref8} directly rendered shadows by integrating geometry and solar position, leveraging UTM coordinates and multi-parameter joint optimization to improve accuracy. On the application side, SpS-NeRF~\cite{ref33} combined traditional MVS~\cite{ref1} depth priors to enhance sparse-view rendering; Sat-Mesh\cite{ref34} used signed distance functions for high-quality surface reconstruction; Season-NeRF~\cite{ref35} introduced temporal encoding for seasonal feature rendering; and Snake-NeRF~\cite{ref36} extended NeRF to large-scale satellite 3D reconstruction. Further efficiency improvements have been achieved by methods like RS-NeRF~\cite{ref37}, SatNGP~\cite{ref38}, and SatensoRF~\cite{ref39} through various acceleration strategies.

\subsection{3DGS for Satellite Images}

3DGS~\cite{ref9} initializes Gaussian scenes from SfM point clouds~\cite{ref40}, explicitly representing scenes with anisotropic 3D Gaussians and rendering via differentiable rasterization. Combining the strengths of neural fields and point-based rendering, 3DGS achieves high-fidelity, real-time rendering and has attracted significant attention. Recent extensions include geometry reconstruction~\cite{ref41,ref42,ref43,ref44}, rendering quality improvement~\cite{ref45, ref46}, sparse-view generalization~\cite{ref47,ref48,ref49}, and inverse rendering/relighting~\cite{ref50,ref51,ref52,ref53}. EVER~\cite{ref54} and 3DGRT~\cite{ref55} further integrated ray tracing into 3DGS, enabling complex camera models and accurate shadow computation.

In satellite imagery, 3DGS has been specialized for remote sensing applications: EOGS~\cite{ref11} first adapted 3DGS to satellite data using an affine camera model for RPC fitting and implemented shadow mapping for shadow rendering; SatGS~\cite{ref10} incorporated appearance embedding and uncertainty modeling to handle seasonal variations and transient objects; Skysplat~\cite{ref56} proposed a feedforward 3DGS framework for rapid reconstruction from sparse multi-temporal images; and Skyfall-GS~\cite{ref57} combined 3DGS with diffusion models and curriculum learning to synthesize navigable 3D cityscapes with geometric consistency and visual realism.

\section{Method}

In this section, we introduce ShadowGS, a novel framework based on 3D Gaussian Splatting, designed to decouple geometric properties (e.g., normal, depth), appearance attributes (e.g., albedo), and illumination components (e.g., direct sunlight, skylight, and near-surface reflection) from multi-temporal satellite image collections. The overall pipeline is illustrated in \cref{fig2}.

\noindent
\textbf{3DGS Basics:} Each Gaussian ellipsoid is parameterized by a center position $\bm{\mu} \in \mathbb{R}^3$, a scaling factor $\bm{\mathbf{s}} \in \mathbb{R}^3$, and a rotation quaternion $\bm{\mathbf{q}} \in \mathbb{R}^4$. Its spatial influence is defined by a covariance matrix $\bm{\Sigma} \in \mathbb{R}^{3 \times 3}$, constructed from a scaling matrix $\mathbf{S}$ (derived from $\mathbf{s}$) and a rotation matrix $\mathbf{R}$ (obtained from $\mathbf{q}$) as $\bm{\Sigma} = \mathbf{RSS}^T\mathbf{R}^T$. The 3D Gaussian distribution $G(\mathbf{x})$ is formulated as:
\begin{equation}
G(\mathbf{x}) = e^{-(\mathbf{x} - \bm{\mu})^T \bm{\Sigma}^{-1} (\mathbf{x} - \bm{\mu})}
\end{equation}

3DGS employs EWA splatting \cite{ref58} to project 3D Gaussians onto the 2D image plane. The projected 2D covariance matrix $\bm{\Sigma}'$ is given by:
\begin{equation}
\bm{\Sigma}' = \mathbf{JW\Sigma W}^T\mathbf{J}^T
\end{equation}
where $\mathbf{W}$ denotes the viewing transformation matrix from world to camera coordinates, and $\mathbf{J}$ is the Jacobian of the projective transformation.

In addition to geometry, each Gaussian stores an opacity value $o$ and a set of learnable spherical harmonics (SH) coefficients that model the view-dependent appearance $c$. The color $C$ of a pixel is computed via alpha blending:
\begin{equation}
C = \sum_i T_i \alpha_i c_i, \quad T_i = \prod_{j=1}^{i-1} (1 - \alpha_j)
\end{equation}
Here, $\alpha_i$ is the pixel translucency of the $i$-th Gaussian, determined by the opacity of the $i$-th Gaussian and the pixel’s position.

During training, all Gaussian parameters are optimized via a photometric loss $\mathcal{L}_c$:
\begin{equation}
\mathcal{L}_c = (1 - \lambda_{\text{ssim}})\mathcal{L}_{\text{color}} + \lambda_{\text{ssim}} \mathcal{L}_{D\text{-SSIM}}
\end{equation}
where $\mathcal{L}_{\text{color}}$ and $\mathcal{L}_{D\text{-SSIM}}$ denote the color reconstruction loss and structural similarity loss, respectively, and $\lambda_{\text{ssim}}$ controls the balance between them.

\subsection{Camera Model and Geometry}
\label{sec:GEO}

\textbf{Camera Model:} The standard 3DGS framework is built on the pinhole camera model, whereas satellite imagery typically adopts the Rational Polynomial Camera (RPC) model to map image coordinates to geographic locations. As shown in \cite{ref59}, the pinhole model introduces only minor error when approximating the RPC model within a local region. Therefore, ShadowGS fits the RPC model using a pinhole camera to align with the existing 3DGS pipeline. Specifically, ShadowGS first refines the original RPC parameters via bundle adjustment \cite{ref7} to generate a sparse point cloud for initializing the 3D Gaussians. A pinhole model is then used in the local tangent plane coordinate system to approximate the optimized RPC model. The average reprojection error for RPC model fitting across all scenes in ShadowGS remains below 0.5 pixels.

\noindent
\textbf{Geometric Representation:} Radiance-based methods often suffer from geometry–radiance ambiguity \cite{ref60}, which complicates the accurate recovery of geometry and appearance in real scenes. Numerous relighting and inverse rendering techniques \cite{ref50,ref51,ref52,ref53} address this by explicitly defining depth and normal attributes in 3DGS and introducing geometric regularization. Following RadeGS \cite{ref43}, we adopt an explicit ray–Gaussian intersection strategy to determine the Gaussian's normal direction and its depth contribution per pixel. Specifically, for a 3D Gaussian $G$, let $(u_c, v_c)$ denote the center of its 2D projection. For a pixel $(u, v)$, the intersection between the camera ray and the Gaussian forms a 1D Gaussian distribution, whose peak defines the ray–Gaussian intersection point. The corresponding depth $d$ represents the depth contribution of Gaussian $G$ to pixel $(u, v)$ and is given by:
\begin{equation}
d = z_c + \frac{z_c}{t_c} \mathbf{m} \begin{pmatrix} u_c - u \\ v_c - v \end{pmatrix}, \quad \hat{\mathbf{m}}= \frac{\mathbf{v}^T \bm{\Sigma}'^{-1}}{\mathbf{v}^T \bm{\Sigma}'^{-1} \mathbf{v}}
\end{equation}
Here, $z_c$ and $t_c$ represent the depth values of the Gaussian center and the distance from the Gaussian center to the camera center, respectively, and the vector $\mathbf{m}$ is a $1 \times 2$ vector formed by omitting the third row of $\hat{\mathbf{m}}$, $\mathbf{v} = (0, 0, 1)^T$. This formulation implies that the intersection between the 3D Gaussian and the camera ray defines a surface, where each pixel corresponds to a different depth value. The normal vector $\mathbf{n}$ of this surface is defined as the Gaussian's normal vector:
\begin{equation}
\mathbf{n} = \mathbf{J}^T (-\mathbf{m} \quad -1)^T
\end{equation}

The normalized depth $D$ and normal maps $\mathbf{N}$ are rendered via alpha blending:
\begin{equation}
\begin{pmatrix} D \\ \mathbf{N} \end{pmatrix} = \sum_i \frac{T_i \alpha_i}{\sum_i T_i \alpha_i} \begin{pmatrix} d_i \\ \mathbf{n}_i \end{pmatrix}
\end{equation}

To further enhance geometric detail, we apply a depth–normal consistency loss $\mathcal{L}_n$ \cite{ref41}:
\begin{equation}
\mathcal{L}_n=(1-\mathbf{N}^T\tilde{\mathbf{N}})
\end{equation}
where $\tilde{\mathbf{N}}$ denotes the surface normal derived from the rendered depth map $D$ via finite differences.

\subsection{Physics-based Rendering Equation}
\label{sec:render}
\textbf{Ray-marching shadow:} Following the hardware-accelerated ray-tracing pipeline for 3DGS introduced in \cite{ref55}, we model shadows using an efficient ray-marching strategy. In satellite scenes, the sun is considered as the sole directional light source. We leverage the ray tracer to evaluate the solar visibility of each Gaussian and combine it with the standard 3DGS rasterizer for pixel-accurate shadow rendering.

Specifically, all Gaussians are organized into a stretched icosahedron bounding volume hierarchy (BVH). The bounding boxes are adaptively scaled according to each Gaussian’s opacity and geometry by applying the following transformation to the icosahedron vertices $\mathbf{a}$ \cite{ref55}:
\begin{equation}
\mathbf{a} \leftarrow \mathbf{a} \sqrt{2 \log(o / o_{\text{min}})} \mathbf{SR}^T + \bm{\mu}
\end{equation}

To ensure the bounding volume fully covers the effective region of each Gaussian, a transparency threshold $o_{\text{min}} = 0.001$ is applied. For each Gaussian center, taken as the ray origin $\bm{\mu}_0$, a ray is cast along the solar direction $\mathbf{r}$. Using a fixed step size, intersections with other Gaussians are detected. The intersection point $\tau$ with the $i$-th Gaussian is defined as the peak of the 1D Gaussian distribution $G^{1D}$ formed by the ray–Gaussian intersection (consistent with Section \ref{sec:GEO}), and is computed as: 
\begin{equation}
\tau = \frac{(\bm{\mu}_0 - \bm{\mu}_i)^T \bm{\Sigma}_i^{-1} \mathbf{r}}{\mathbf{r}^T \bm{\Sigma}_i^{-1} \mathbf{r}}
\end{equation}

The response value $\tilde{\alpha}$ of the intersecting Gaussian along the 
ray is:
\begin{equation}
\tilde{\alpha} = o G^{1D}(\tau)
\end{equation}

The solar visibility $S_{\text{sun}}$ of the current Gaussian is then given by:
\begin{equation}
S_{\text{sun}} = \prod_{i=1}^k (1 - \tilde{\alpha}_i)
\end{equation}
where $k$ is the total number of intersecting Gaussians and $\tilde{\alpha}_i$ is the response value of the $k$-th intersected Gaussian.

\noindent
\textbf{Remote Sensing Physics-based Rendering Equation:} To effectively decouple illumination components and appearance attributes in satellite imagery, we model skylight using a set of globally shared spherical harmonics (SH). This representation captures spatially uniform skylight radiance $l_{\text{sky}}$, with higher-order SH terms disabled to restrict learning to low-frequency features. Each Gaussian is assigned two sets of SH coefficients: one encoding albedo $f$, and the other representing reflected radiance $l_n$ from nearby surfaces. This allows each primitive to model independent material properties and local light interactions. As in standard 3DGS, we progressively enable higher-order SH terms to represent high-frequency details in appearance and near-surface reflections. We render albedo and illumination component maps via alpha blending:
\begin{equation}
\begin{pmatrix} S \\ L_{\text{sky}} \\ L_n \\ F \end{pmatrix} = \sum_{i} T_i \alpha_i \begin{pmatrix} S_{\text{sun}} \\ l_{\text{sky}} \\ l_n \\ f \end{pmatrix}_i
\end{equation}

The total incident radiance at a surface point is computed as:
\begin{equation}
L_{\text{total}} = S + (1 - S) \cdot (L_{\text{sky}} + L_n)
\end{equation}

This formulation ensures that sunlit regions are dominated by direct solar radiance $S$, while shadowed areas are illuminated by skylight $L_{\text{sky}}$ and reflections $L_n$ from nearby surfaces. The final rendered color is obtained as:
\begin{equation}
C = F \cdot L_{\text{total}}
\end{equation}

\subsection{Shadow Consistency Constraint}

The discrete 3D Gaussian representation in 3DGS exhibits greater irregularity compared to NeRF's continuous neural representation, often leading to insufficient optimization constraints. To address this limitation, we introduce a shadow consistency constraint that leverages unique shadow formation characteristics in satellite imaging.

\begin{figure}[htb]
	\centering
	\includegraphics[width=\linewidth]{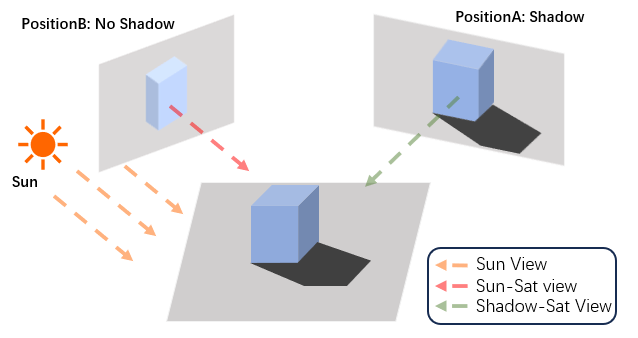}
	\caption{Shadow consistency constraint. Shadows are visible when the satellite's viewing direction differs from the solar direction (position A), but become self-occluded by the object's geometry when the two directions align (position B).}
	\label{fig3}
\end{figure}

As shown in \cref{fig3}, when the satellite is at position A, shadows cast by objects under sunlight are visible in the captured image. However, at position B where the satellite view direction aligns with the sun direction, these shadows become self-occluded by the object's own geometry, resulting in shadow-free imaging. This phenomenon occurs due to the parallel light characteristics of both solar illumination and satellite viewing rays. We formalize this constraint as follows. Given a virtual camera viewpoint and a collinear sun direction, we render the corresponding shadow map $S_v$ using the method in Section \ref{sec:render}. The shadow consistency loss is defined as:
\begin{equation}
\mathcal{L}_{S_1} = \| S_v - \mathbf{1} \|_1
\end{equation}

In practice, we implement this constraint under two configurations: (1) fixing the camera viewpoint while aligning the sun direction with the view direction, and (2) simultaneously adjusting both sun and camera directions to be perpendicular to the scene surface. In both cases, the constraint drives the rendered shadow map toward a fully illuminated state under collinear light-view conditions. This constraint encourages surface Gaussians to achieve higher opacity and promotes better alignment with the underlying geometric surfaces, thereby enhancing reconstruction quality.

\subsection{Shadow Prior for Sparse View}

Sparse input views pose significant challenges for both 3DGS and NeRF frameworks, often leading to overfitting and degraded performance. Existing approaches typically rely on depth priors or diffusion models \cite{ref47,ref48,ref49, ref64}, yet these face limitations: depth-based methods struggle in textureless regions, while diffusion-based techniques may introduce erroneous pseudo-views.

ShadowGS addresses this by leveraging geometry-correlated shadow information as a robust supervision signal \cite{ref32}. We integrate FDRNet \cite{ref16}, a self-supervised shadow detection network that demonstrates low false-negative rates in satellite imagery. Under sparse-view conditions, FDRNet-provided shadow priors effectively guide the optimization of global geometry and illumination parameters, promoting stable convergence.

Specifically, we minimize the discrepancy between the rendered shadow map $S$ and the FDRNet-extracted shadow mask $\hat{S}$ using a binary cross-entropy loss:
\begin{equation}
\mathcal{L}_{S_3} = -(S \log_2(\hat{S}) + (1 - S) \log_2(1 -\hat{S} ))
\end{equation}
 
To address false positives in vegetation areas, we employ the Normalized Difference Vegetation Index (NDVI) with multi-spectral data, or the Difference Enhanced Vegetation Index (DEVI) for RGB imagery, excluding detected vegetation regions from shadow supervision. Furthermore, we discontinue the shadow prior loss after densification concludes to prevent interference from remaining false positives.

\begin{table*}[htbp]
\centering
\resizebox{\linewidth}{!}{\begin{tabular}{
l *{7}{Y} Y *{3}{Y} Y
}
\toprule
\multirow{5}{*}{\textbf{AOI}} 
  & \multicolumn{14}{c}{\textbf{DFC2019 (RGB)}} 
  & \multicolumn{10}{c}{\textbf{IARPA(Pansharpened)}} \\
  & \multicolumn{14}{c}{\textbf{PSNR(dB)$\uparrow$ / MAE(m)$\downarrow$}}
  & \multicolumn{10}{c}{\textbf{PSNR(dB)$\uparrow$ / MAE(m)$\downarrow$}} \\
  & \multicolumn{2}{c}{JAX\_004} & \multicolumn{2}{c}{JAX\_068} & \multicolumn{2}{c}{JAX\_165}
  & \multicolumn{2}{c}{JAX\_214} & \multicolumn{2}{c}{JAX\_260} & \multicolumn{2}{c}{OMA\_288}
  & \multicolumn{2}{c}{OMA\_315} & \multicolumn{2}{c}{\textbf{Mean}}
  & \multicolumn{2}{c}{IARPA\_001} & \multicolumn{2}{c}{IARPA\_002} & \multicolumn{2}{c}{IARPA\_003}
  & \multicolumn{2}{c}{\textbf{Mean}} \\
\midrule

S2P\cite{ref62}
& {--}&{2.88} & {--}&{1.64} & {--}&{4.45} & {--}&{2.82} & {--}&{2.02}
& {--}&{3.21} & {--}&{1.97} & {--}&{2.71} 
& {--}&{3.00} & {--}&{4.65} & {--}&{2.37} & {--}&{3.34} \\

Sat-NGP\cite{ref38}
& {20.33}&{1.47} & {18.14}&{1.43} & {23.14}&{2.20} & {17.99}&{1.99} & {18.73}&{2.25}
& {15.38}&{5.08} & {14.99}&{\textbf{1.72}} & {18.39}&{2.31} 
& {--}&{--} & {--}&{--} & {--}&{--} & {--}&{--} \\

EO-NeRF\cite{ref8}
& {20.16}&{\textbf{1.41}} & {17.80}&{1.56} & {23.46}&{3.39} & {16.76}&{2.76} & {18.57}&{1.82}
& {17.79}&{4.87} & {15.51}&{1.79} & {18.58}&{2.51} 
& {22.17}&{1.56} & {21.06}&{\textbf{1.87}} & {19.17}&{2.34} & {20.80}&{1.92} \\

EOGS\cite{ref11}
& {22.56}&{2.06} & {21.44}&{2.10} & {17.88}&{4.21} & {18.54}&{3.62} & {20.60}&{3.90}
& {15.43}&{19.45} & {15.84}&{6.22} & {18.90}&{5.94} 
& {22.35}&{1.58} & {23.43}&{1.99} & {\textbf{24.65}}&{\textbf{1.28}} & {23.48}&{1.62} \\

\textbf{Ours}
& {\textbf{24.11}}&{1.61} & {\textbf{24.10}}&{\textbf{0.98}} & {\textbf{25.20}}&{\textbf{1.55}} & {\textbf{22.20}}&{\textbf{1.58}} & {\textbf{23.04}}&{\textbf{1.32}}
& {\textbf{17.92}}&{\textbf{2.79}} & {\textbf{16.23}}&{1.99} & {\textbf{21.83}}&{\textbf{1.69}} 
& {\textbf{24.95}}&{\textbf{1.44}} & {\textbf{24.30}}&{1.85} & {24.74}&{1.52} & {\textbf{24.66}}&{\textbf{1.60}} \\

\bottomrule
\end{tabular}}
\caption{Quantitative comparison of novel view synthesis and 3D reconstruction results across 10 AOIs under multi-view input. Best results are bolded.}
\label{tab:1}
\end{table*}

\begin{table*}[htbp]
\centering
\resizebox{\linewidth}{!}{
\begin{tabular}{
l *{7}{Y} 
Y
}
\toprule
\multirow[b]{2}{*}{\textbf{AOI}} 
  & \multicolumn{14}{c}{\textbf{BER(\%)$\downarrow$  / ACC(\%)$\uparrow$ }}
  & \multicolumn{2}{c}{\multirow[b]{2}{*}{\textbf{Mean}}} \\
  & \multicolumn{2}{c}{JAX\_004} & \multicolumn{2}{c}{JAX\_068} & \multicolumn{2}{c}{JAX\_165}
  & \multicolumn{2}{c}{JAX\_214} & \multicolumn{2}{c}{JAX\_260} & \multicolumn{2}{c}{OMA\_288}
  & \multicolumn{2}{c}{OMA\_315} \\
\midrule

S-EO\cite{ref32}
& {48.11}&{85.55} & {22.98}&{92.92} & {--}&{--} & {22.41}&{89.54} & {40.52}&{86.06}
& {--}&{--} & {--}&{--} & {--}&{--} \\

FSDNet\cite{ref26}
& {41.81}&{83.15} & {26.67}&{90.90} & {16.29}&{89.15} & {23.79}&{86.53} & {41.16}&{72.73}
& {31.83}&{79.95} & {29.08}&{91.88} & {30.09}&{84.90} \\

FDRNet\cite{ref16}
& {31.97}&{78.04} & {20.81}&{91.20} & {15.09}&{85.92} & {18.55}&{84.84} & {32.10}&{82.07}
& {29.42}&{80.50} & {26.07}&{90.73} & {24.86}&{84.76} \\
EO-NeRF\cite{ref8}
& {25.50}&{86.67} & {26.55}&{91.91} & {40.97}&{79.59} & {38.58}&{83.08} & {33.80}&{87.59}
& {40.31}&{79.19} & {49.88}&{88.92} & {36.51}&{85.28} \\
EOGS\cite{ref11}
& {35.68}&{\textbf{88.19}} & {15.88}&{\textbf{92.76}} & {17.44}&{89.24} & {19.80}&{88.88} & {28.68}&{\textbf{88.32}}
& {51.19}&{72.20} & {47.81}&{89.16} & {30.93}&{86.97} \\
\textbf{Ours}
& {\textbf{22.43}}&{86.54} & {\textbf{12.37}}&{91.95} & {\textbf{10.26}}&{\textbf{91.23}} & {\textbf{11.84}}&{\textbf{90.72}} & {\textbf{18.89}}&{87.88}
& {\textbf{24.92}}&{\textbf{85.39}} & {\textbf{24.09}}&{\textbf{92.19}} & {\textbf{17.83}}&{\textbf{89.41}} \\

\bottomrule
\end{tabular}}
\caption{Quantitative comparison of shadow detection performance across 7 AOIs from DFC2019 under multi-view input. Best results are shown in bold.}
\label{tab:2}
\end{table*}

\subsection{Training Strategy and Total Loss}

\begin{figure}[htb]
	\centering
	\includegraphics[width=\linewidth]{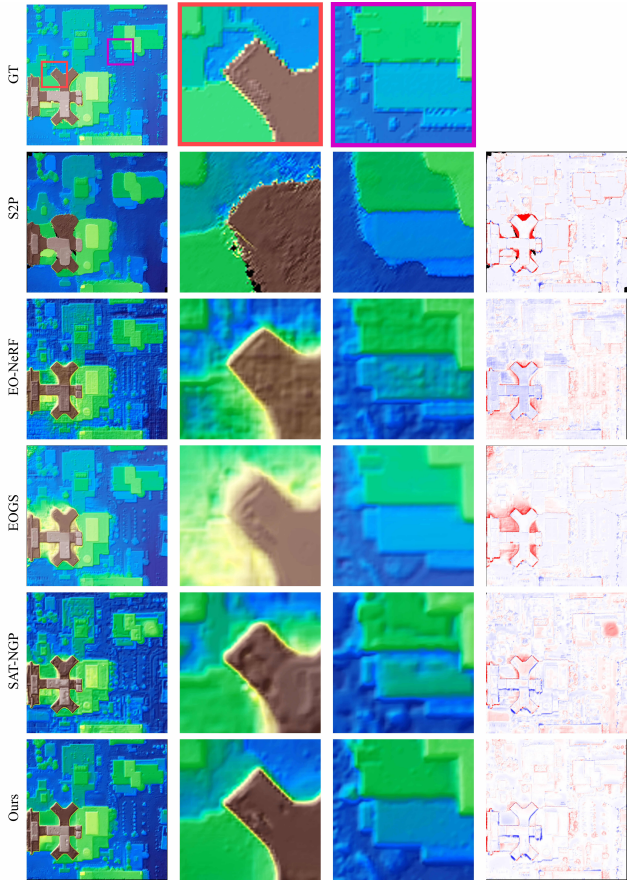}
	\caption{Geometric reconstruction visualization on the JAX 068 dataset. The fourth column shows the error map between each method's reconstructed DSM and the ground truth (GT), where red indicates overestimation and blue indicates underestimation of height values.}
	\label{fig4}
\end{figure}

\textbf{Total Loss:} In addition to the aforementioned losses, we incorporate a binary cross-entropy loss $\mathcal{L}_{S_2}$ —consistent with EOGS \cite{ref11}—to encourage projected shadows to converge toward binary states (fully transparent or opaque):
\begin{equation}
\mathcal{L}_{S_2} = -(S \log_2(S) + (1 - S) \log_2(1 - S))
\end{equation}

The complete loss function is defined as:
\begin{equation}
\mathcal{L} = \lambda_c \mathcal{L}_c + \lambda_n \mathcal{L}_n + \lambda_{S_1} \mathcal{L}_{S_1} + \lambda_{S_2} \mathcal{L}_{S_2} + \lambda_{S_3} \mathcal{L}_{S_3}
\end{equation}
Here, $\lambda$ are experimentally determined weighting coefficients. For the DFC2019 dataset \cite{ref12,ref13}, we set $\lambda_c = 10$, $\lambda_n = 0.5$, $\lambda_{S_1} = 0.2$, $\lambda_{S_2} = 0.3$, $\lambda_{S_3} = 1$. For IARPA \cite{ref61}, we reduce $\lambda_n$ to 0.1 for optimal performance. The shadow prior loss $\mathcal{L}_{S_3}$ is activated only under sparse-view conditions.

\noindent
\textbf{Training Strategy:} We adapt the original 3DGS training procedure to accommodate ShadowGS's extended parameter set and optimization objectives. Key modifications include:

\begin{figure}[htb]
	\centering
	\includegraphics[width=\linewidth]{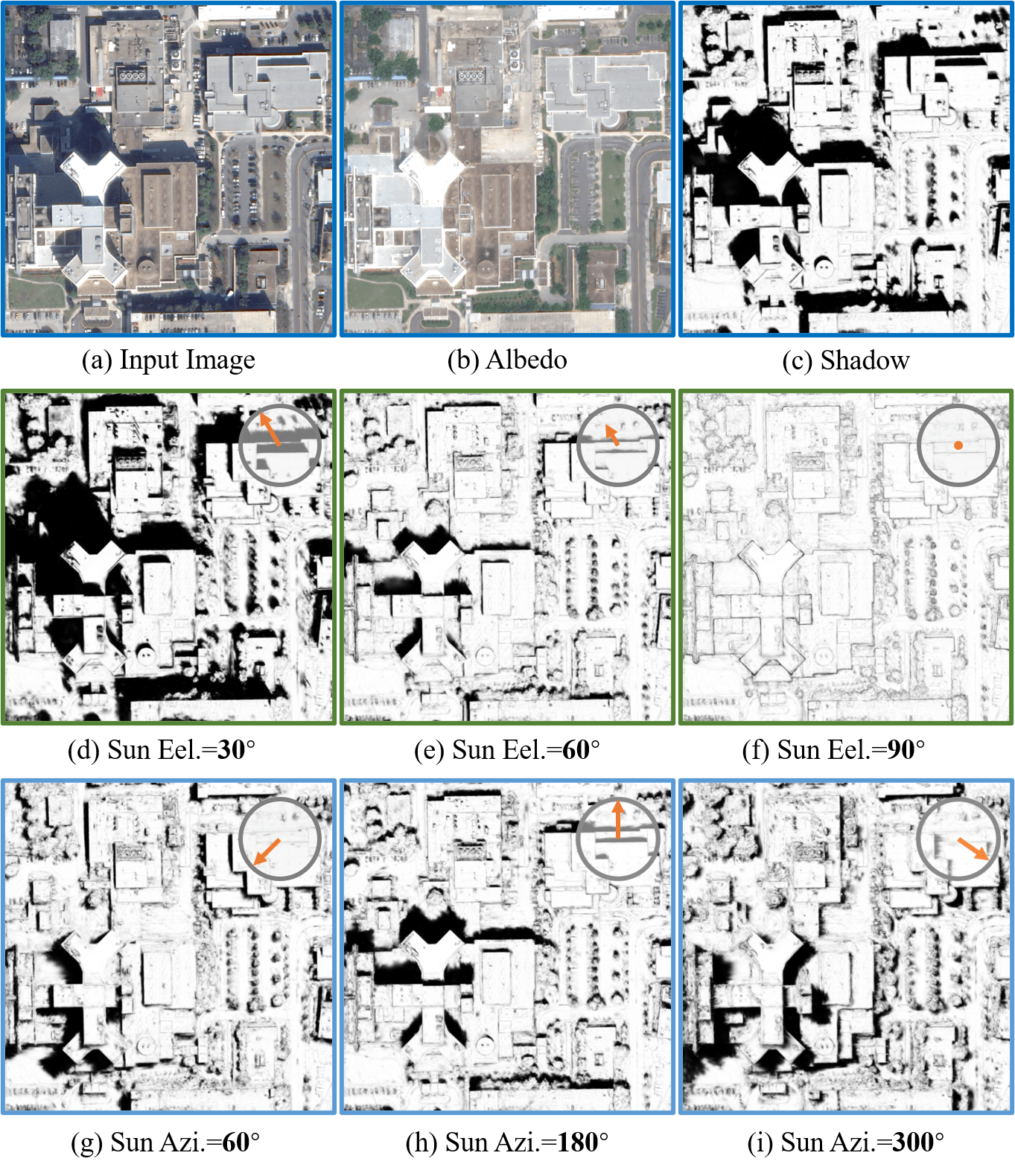}
	\caption{Albedo Decomposition and Shadow Modeling Results. (a) Input image; (b-c) Albedo and shadow maps decoupled by ShadowGS; (d-f) Shadow maps under different solar elevation angles; (g-i) Shadow maps under different solar azimuth angles.}
	\label{fig5}
\end{figure}
\begin{itemize}
    \item Disabling opacity reset throughout training
    \item Activating all loss terms from the first iteration
    \item Setting total iterations to 5000, with densification halted after iteration 3000
    \item Performing densification every 300 iterations during the densification phase
\end{itemize}

These adjustments ensure stable convergence given the increased number of appearance and illumination parameters. Reducing the frequency of densification allows Gaussian primitives to be more fully optimized before further densification occurs. This strategy also helps prevent the proliferation of redundant Gaussians, thereby alleviating computational overhead. 
\section{Experiment}

\textbf{Datasets:} We train and evaluate ShadowGS on datasets from the 2019 IEEE GRSS Data Fusion Competition (DFC2019)~\cite{ref12,ref13} and the 2016 IARPA Multi-View Stereo 3D Mapping Challenge (IARPA2016)~\cite{ref61}. The DFC2019 dataset covers two urban regions—Jacksonville (JAX) and Omaha (OMA)—using RGB imagery, whereas the IARPA2016 dataset includes a single urban area, Buenos Aires, and provides pan-sharpened imagery. All images were captured by the WorldView-3 satellite at a ground sampling distance of 0.3 m/pixel and are provided with RPC parameters, acquisition timestamps, and solar angles. LiDAR-derived Digital Surface Models (DSMs) are available as reference, with resolutions of 0.5 m for DFC2019 and 0.3 m for IARPA2016. Additionally, using solar angles, RPCs, and LiDAR DSMs, we generate accurate shadow masks for DFC2019 following SEO~\cite{ref32} to enable quantitative evaluation of shadow disentanglement.
 
\noindent
\textbf{Experimental Setup:} We evaluate under two input settings: multi-view and sparse-view.
\begin{itemize}
    \item Multi-View Input: We use 10 areas of interest (AOIs): 5 from JAX with diverse terrain features, 2 from OMA with notable seasonal variations, and 3 from IARPA based on pan-sharpened images. Each AOI contains 10–40 multi-temporal images with varying viewpoints. Approximately 15\% of images are held out as the test set, with the rest used for training.
    \item Sparse-View Input: We use 5 AOIs from JAX, each containing 4 images captured under different viewing angles, times, and illumination. Three images are used as input, and the remaining one is reserved for testing.
\end{itemize}

\noindent
\textbf{Evaluation Metrics:} We adopt the following metrics:
\begin{itemize}
    \item Novel View Synthesis: Peak Signal-to-Noise Ratio (PSNR) between rendered and real images.
    \item 3D Reconstruction: Height Mean Absolute Error (MAE) between the reconstructed DSM and the LiDAR reference DSM.
    \item Shadow Disentanglement: Balanced Error Rate (BER) and Accuracy (ACC) between the binarized rendered shadow map and the reference shadow mask.
\end{itemize}
\textbf{Implementation Details:} All experiments are conducted on a Ubuntu 22.04 server with a single NVIDIA RTX 4090 GPU (24 GB). Our implementation builds on the public release of RadeGS~\cite{ref43}. Training time for each AOI is approximately 10 minutes. For the classic multi-view stereo pipeline S2P~\cite{ref62, ref63}, in the multi-view experiments, we generated the DSM following the method provided by SatNeRF~\cite{ref7}; under sparse-view conditions, we generated the DSM by fusing pairwise image matching results.

\subsection{Multi-View Experimental Results}

\cref{tab:1} provides a quantitative comparison between ShadowGS and state-of-the-art methods under multi-view input. ShadowGS achieves higher DSM accuracy across most AOIs and consistently outperforms existing approaches in novel view synthesis. On the DFC2019 dataset, ShadowGS reduces the average height MAE by approximately 0.62 meters and improves the average PSNR by about 2.93 dB compared to the previous best method. On the IARPA dataset, it attains a similar height MAE to EOGS~\cite{ref11} while increasing PSNR by roughly 1.18 dB.

For shadow disentanglement, \cref{tab:2} reports quantitative results on 7 AOIs from DFC2019. ShadowGS reduces the balanced error rate by nearly half while also improving detection accuracy. We also compare against two natural-scene shadow detection methods—FDRNet~\cite{ref16} and FSDNet~\cite{ref26}—as well as the SEO~\cite{ref32} shadow detection network. ShadowGS outperforms all three in shadow detection performance.

\begin{table*}[htbp]
\centering\tabcolsep 0.34cm
\begin{tabular}{
l *{5}{Y} 
Y}
\toprule
\multirow[b]{2}{*}{\textbf{AOI}} 
  & \multicolumn{10}{c}{\textbf{PSNR(dB)$\uparrow$ / MAE(m)$\downarrow$}}
  & \multicolumn{2}{c}{\multirow[b]{2}{*}{\textbf{Mean}}} \\ 
  & \multicolumn{2}{c}{JAX\_004} & \multicolumn{2}{c}{JAX\_068} & \multicolumn{2}{c}{JAX\_165}
  & \multicolumn{2}{c}{JAX\_214} & \multicolumn{2}{c}{JAX\_260}
   \\
\midrule

S2P\cite{ref62}
& {--}&{3.14} & {--}&{\textbf{1.58}} & {--}&{5.47} & {--}&{\textbf{3.42}} & {--}&{3.20}
& {--}&{3.36}  \\

EO-NeRF\cite{ref8}
& {20.83}&{\textbf{1.51}} & {18.24}&{5.81} & {12.53}&{9.85} & {11.93}&{8.40} & {17.44}&{3.38}
& {16.19}&{5.79}  \\

EOGS\cite{ref11}
& {\textbf{22.46}}&{2.64} & {14.02}&{6.63} & {10.11}&{12.90} & {7.13}&{17.40} & {13.33}&{4.98}
& {13.41}&{8.91}  \\

\textbf{Ours}
& {{21.91}}&{3.24} & {18.64}&{1.89} & {18.68}&{3.97} & {17.90}&{4.98} & {20.10}&{2.44}
& {19.45}&{3.30}  \\

\textbf{Ours} + Shadow
& {22.30}&{2.08} & {\textbf{18.66}}&{1.69} & {\textbf{19.10}}&{\textbf{3.36}} & {\textbf{18.25}}&{3.94} & {\textbf{20.27}}&{\textbf{2.26}}
& {\textbf{19.72}}&{\textbf{2.67}}\\
\bottomrule
\end{tabular}
\caption{Quantitative comparison on 5 JAX AOIs under sparse-view input. Best results are shown in bold. "Ours + Shadow" indicates supervision with shadow map from FDRNet\cite{ref16}. }
\label{tab:3}
\end{table*}

\begin{table*}[ht]
  \centering\tabcolsep 0.6cm
  \begin{tabular}{c c c | S[table-format=1.3] S[table-format=1.3] S[table-format=1.3]}
    \toprule
    Depth-Normal & Render Equation & Shadow Consistency & {MAE(m) $\downarrow$} & {PSNR(dB) $\uparrow$}  \\
    \midrule
      &&& 4.00 & 13.91  \\
    \checkmark & & & 2.50 & 17.07  \\
    \checkmark & \checkmark & & 2.11 & 23.00  \\
    \checkmark & \checkmark & \checkmark & \bfseries 1.41 & \bfseries 23.73  \\
    \bottomrule
  \end{tabular}
   \caption{Ablation study on 5 JAX AOIs. Results are averaged over all 5 AOIs.}
  \label{tab:4}
\end{table*}

\cref{fig4} visualizes geometric reconstruction results. ShadowGS reconstructs sharper edges for structural objects such as buildings and yields smoother surfaces in low-texture regions. The overall geometry aligns more closely with LiDAR reference data. Additional reconstruction visualizations are included in the Supplementary material.

\cref{fig5} illustrates ShadowGS’s ability to recover albedo and model geometry-aware shadows. ShadowGS  successfully restores albedo information in shadowed regions and renders shadows consistent with scene geometry under varying sun angles.

\subsection{Sparse-View Experimental Results}

\cref{tab:3} compares ShadowGS with existing methods under sparse-view input. Without using shadow priors, ShadowGS already significantly outperforms EO-NeRF~\cite{ref8} and EOGS~\cite{ref11} in both height MAE and novel view PSNR across all AOIs, and slightly surpasses the classic MVS pipeline S2P~\cite{ref62,ref63} in reconstruction quality. When shadow map priors are incorporated, ShadowGS further improves DSM accuracy and synthesis performance under sparse views. 

\subsection{Ablation Studies}

We conduct ablation experiments to evaluate the contribution of each major component in ShadowGS to geometric reconstruction and novel view synthesis. \cref{tab:4} summarizes the results, where each row corresponds to a different model configuration and columns indicate the activation of the following components: depth–normal consistency constraint, rendering equation, and shadow consistency constraint.

Experiments are conducted under multi-view settings, with the last two columns reporting the average height MAE and PSNR across five JAX AOIs. Results show that adding the depth–normal constraint reduces MAE by 1.50 meters and increases PSNR by 3.16 dB. Enabling the rendering equation further improves PSNR by 5.93 dB and reduces MAE by 0.39 meters. Incorporating the shadow consistency constraint brings an additional MAE reduction of 0.70 meters and a PSNR gain of 0.73 dB. In summary, all components contribute positively to both reconstruction and rendering quality.
\section{Conclusion}

We propose ShadowGS, a novel 3DGS-based framework that decouples geometry, appearance, and illumination from multi-temporal satellite images—including RGB, pan-sharpened, and sparse-view inputs. By introducing a remote-sensing physics-based rendering equation combined with efficient ray marching, ShadowGS accurately models shadow variations across multi-temporal observations, effectively disentangles illumination and appearance, and achieves high-quality geometric reconstruction.

A limitation of ShadowGS is that it does not currently account for content inconsistencies in multi-temporal imagery caused by seasonal or land-cover changes, which may affect performance in dynamically varying scenes. Future work could incorporate seasonal appearance modeling into the rendering equation to improve robustness in such scenarios.
\newpage
{
    \small
    \bibliographystyle{ieeenat_fullname}
    \bibliography{main}
    \nocite{*}
}

\clearpage
\maketitlesupplementary

\section*{A. Additional Experimental Results under Multi-view Input}

\subsection*{A.1. Geometric Reconstruction Visualization}
We provide additional 3D reconstruction visualizations under multi-view input in Figures 6 to 14.

\begin{figure}[ht]
    \centering
    \includegraphics[width=\linewidth]{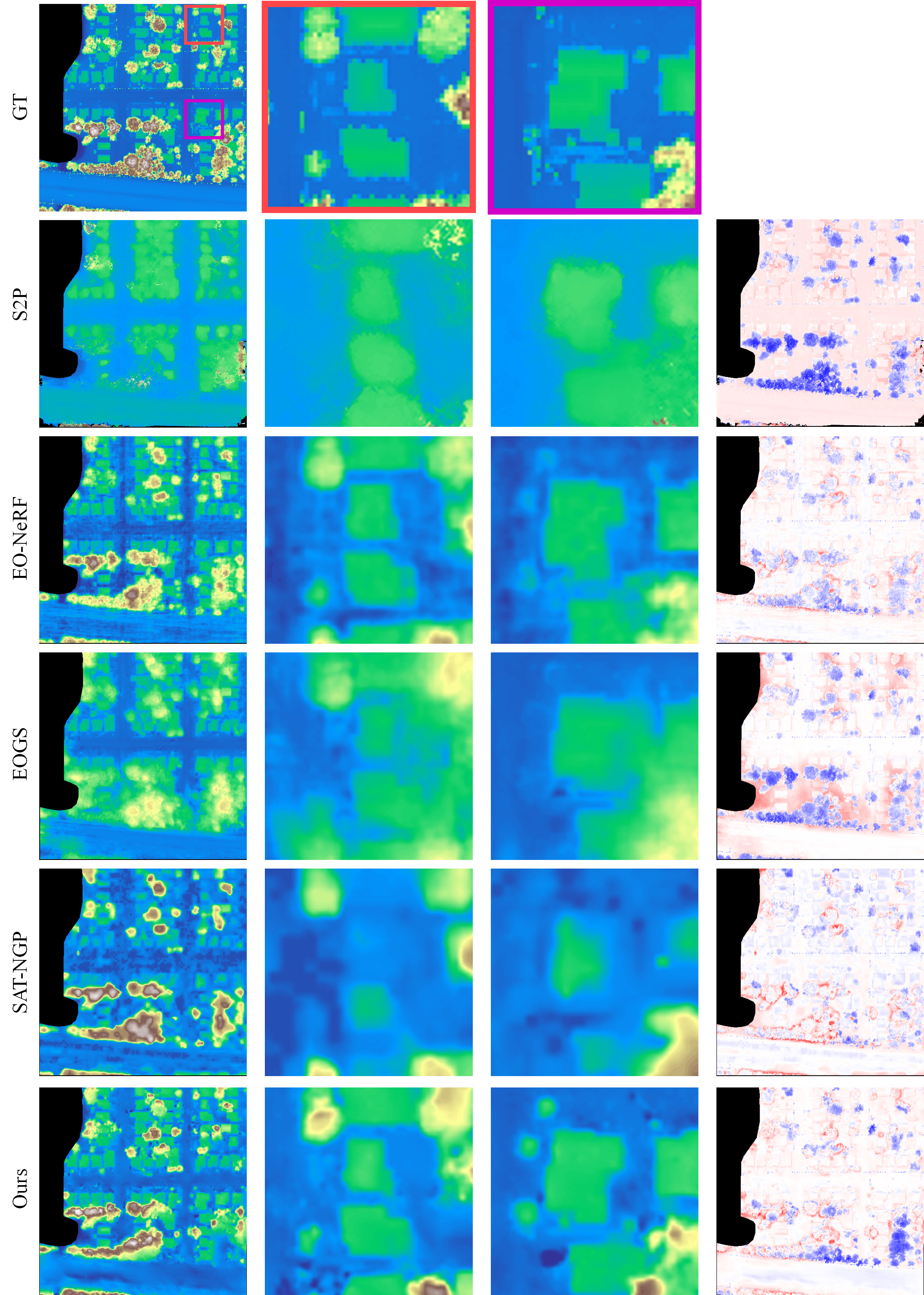}
    \caption{Geometric reconstruction visualization on the JAX 004 dataset. The fourth column shows the error map between each method's reconstructed DSM and the ground truth (GT), where red indicates overestimation and blue indicates underestimation of height values.}
    \label{fig:6}
\end{figure}
\newpage

\begin{figure}[ht]
    \centering
 \includegraphics[width=\linewidth]{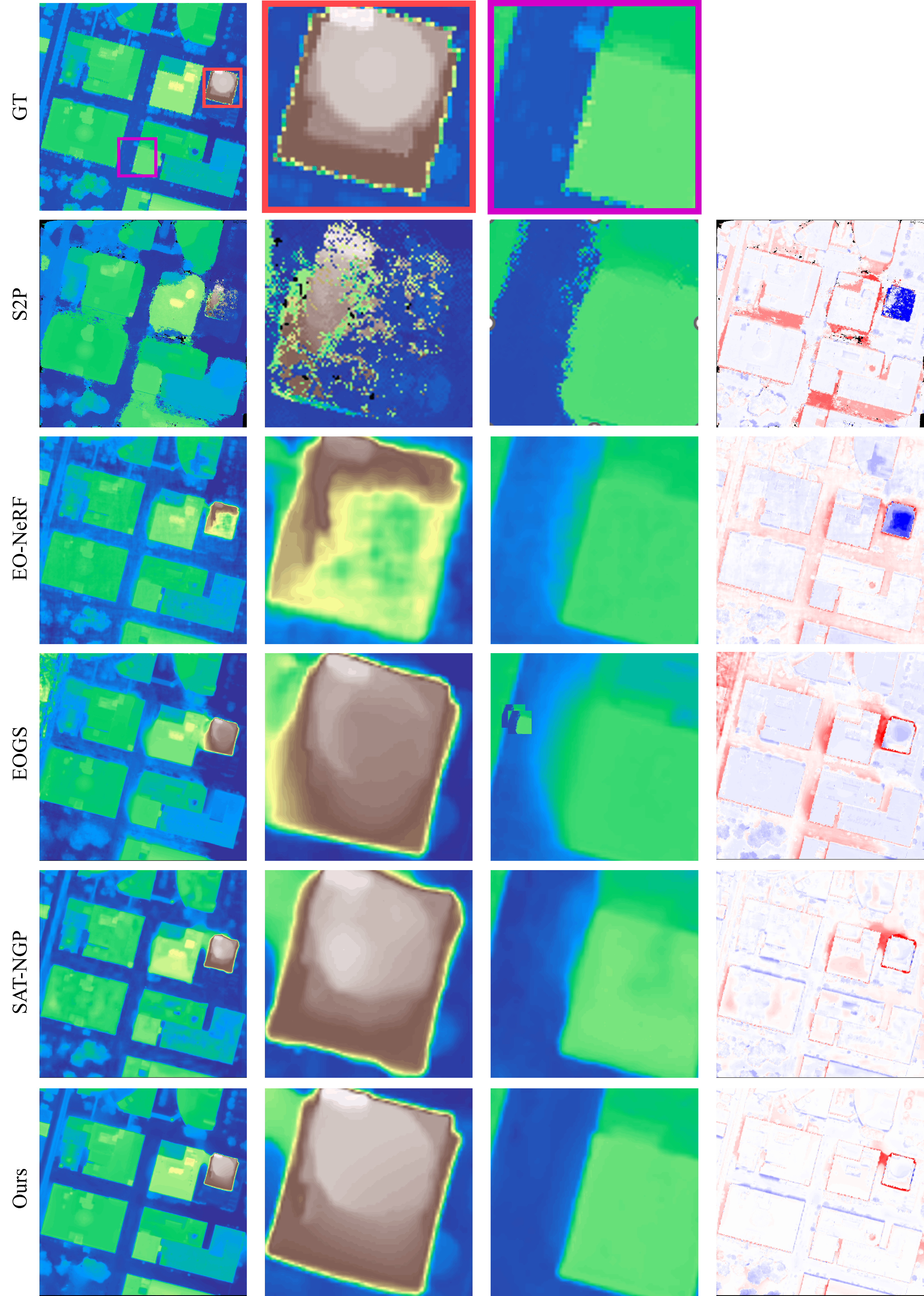}
    \caption{Geometric reconstruction visualization on the JAX 165 dataset. The fourth column shows the error map between each method's reconstructed DSM and the ground truth (GT), where red indicates overestimation and blue indicates underestimation of height values.}
    \label{fig:76}
\end{figure}
\newpage

\begin{figure}[ht]
    \centering
 \includegraphics[width=\linewidth]{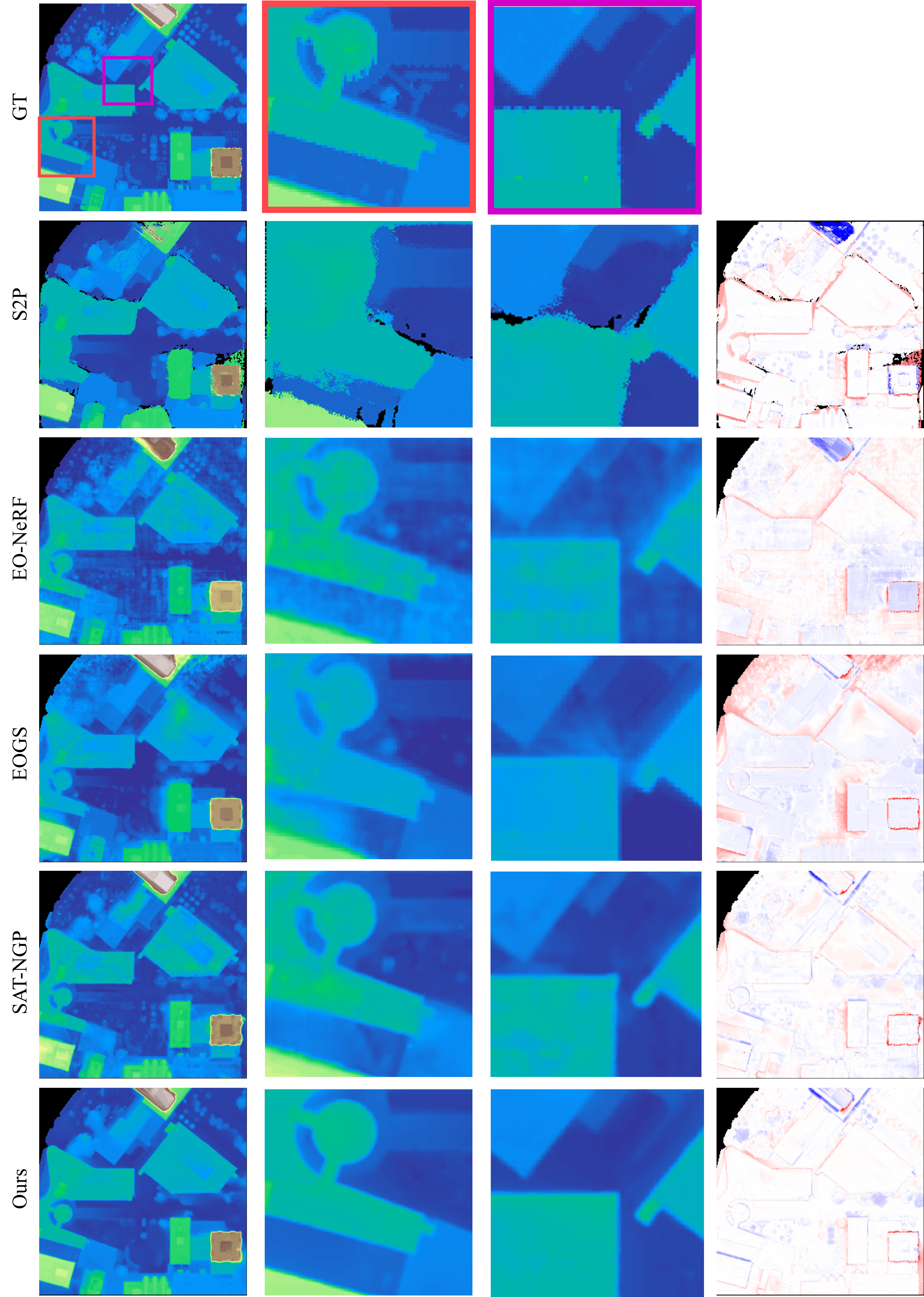}
    \caption{Geometric reconstruction visualization on the JAX 214 dataset. The fourth column shows the error map between each method's reconstructed DSM and the ground truth (GT), where red indicates overestimation and blue indicates underestimation of height values.}
    \label{fig:8}
\end{figure}
\newpage

\begin{figure}[ht]
    \centering
 \includegraphics[width=\linewidth]{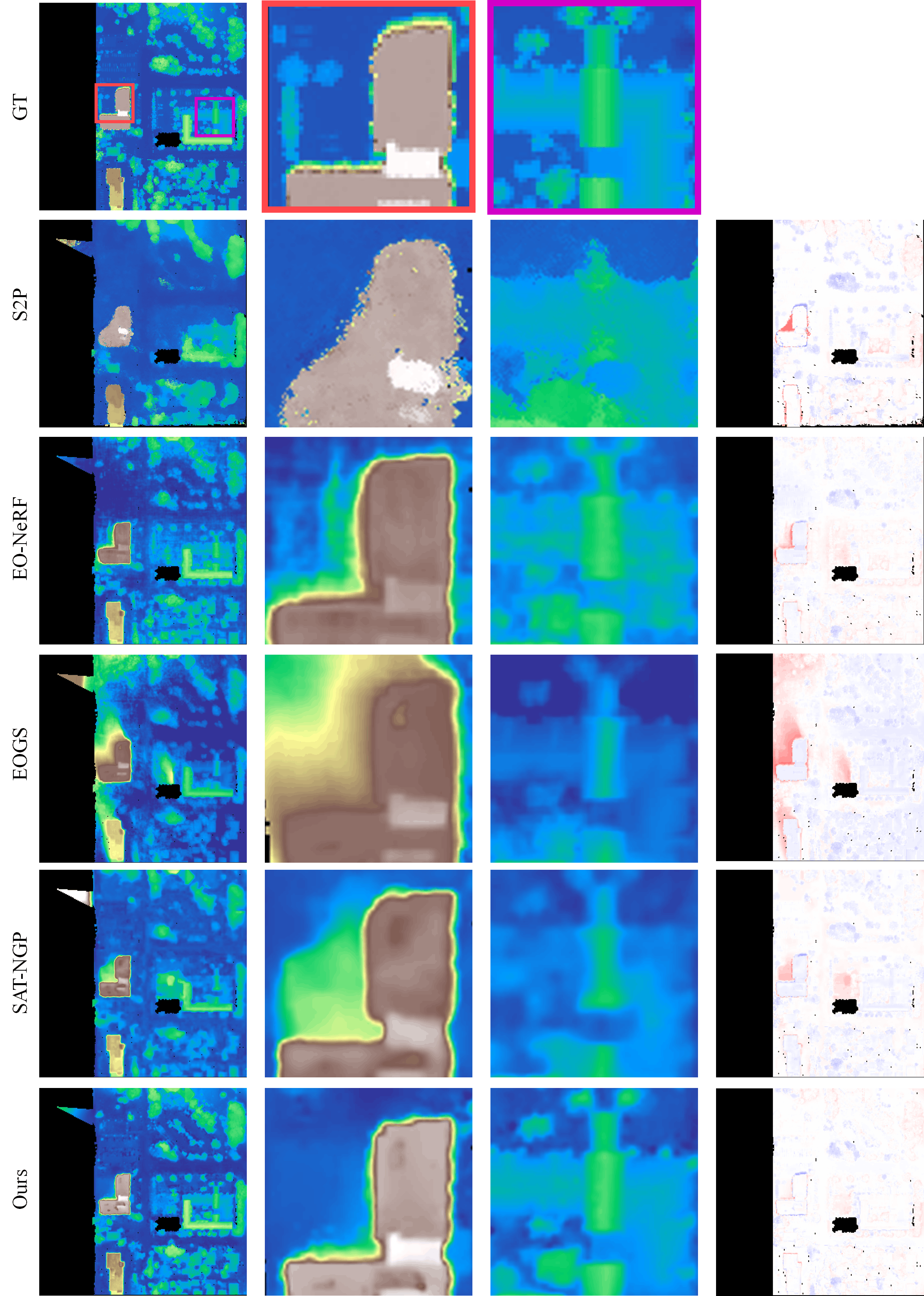}
    \caption{Geometric reconstruction visualization on the JAX 260 dataset. The fourth column shows the error map between each method's reconstructed DSM and the ground truth (GT), where red indicates overestimation and blue indicates underestimation of height values.}
    \label{fig:9}
\end{figure}
\newpage

\begin{figure}[ht]
    \centering
 \includegraphics[width=\linewidth]{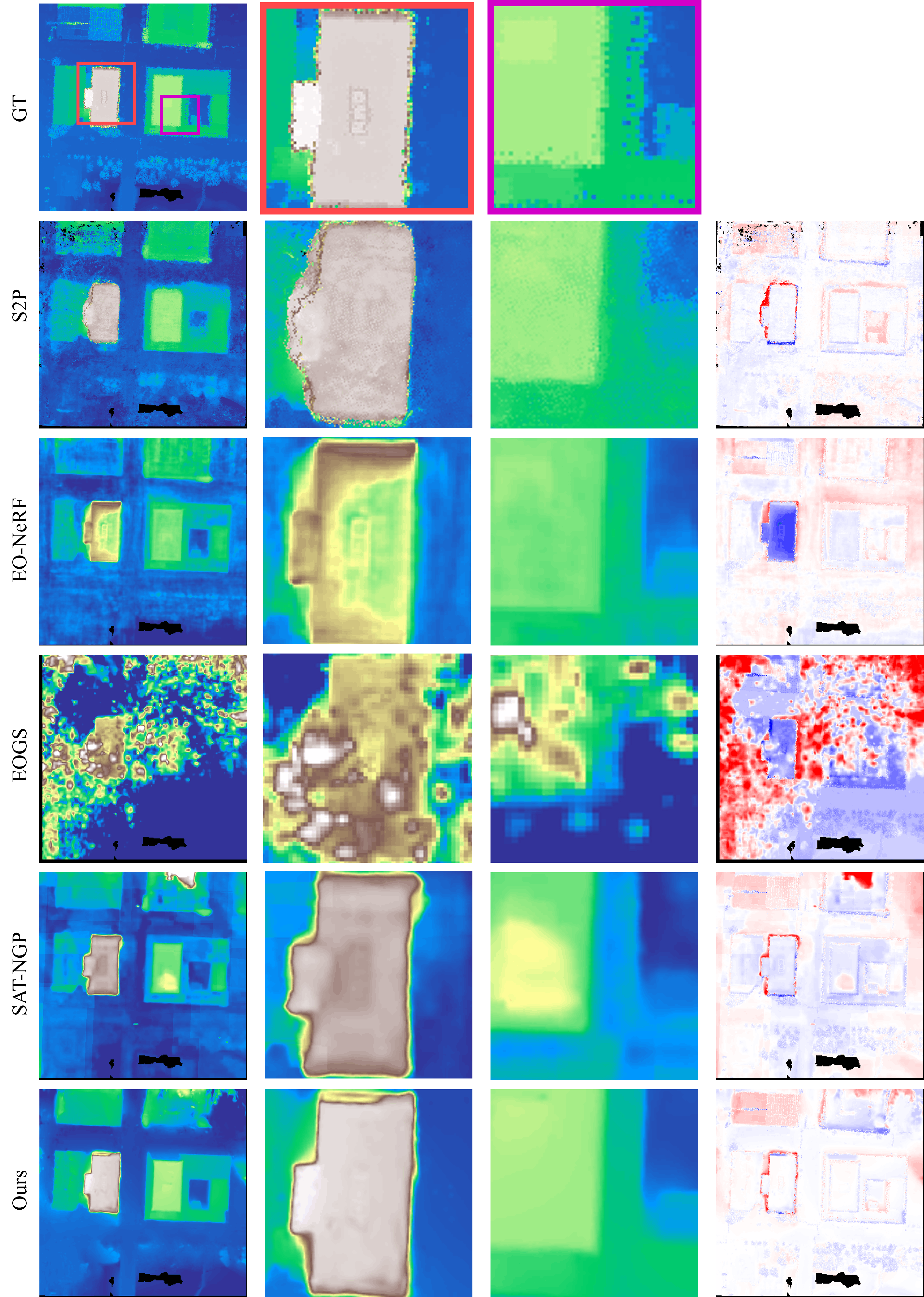}
    \caption{Geometric reconstruction visualization on the OMA 288 dataset. The fourth column shows the error map between each method's reconstructed DSM and the ground truth (GT), where red indicates overestimation and blue indicates underestimation of height values.}
    \label{fig:10}
\end{figure}
\newpage

\begin{figure}[ht]
    \centering
 \includegraphics[width=\linewidth]{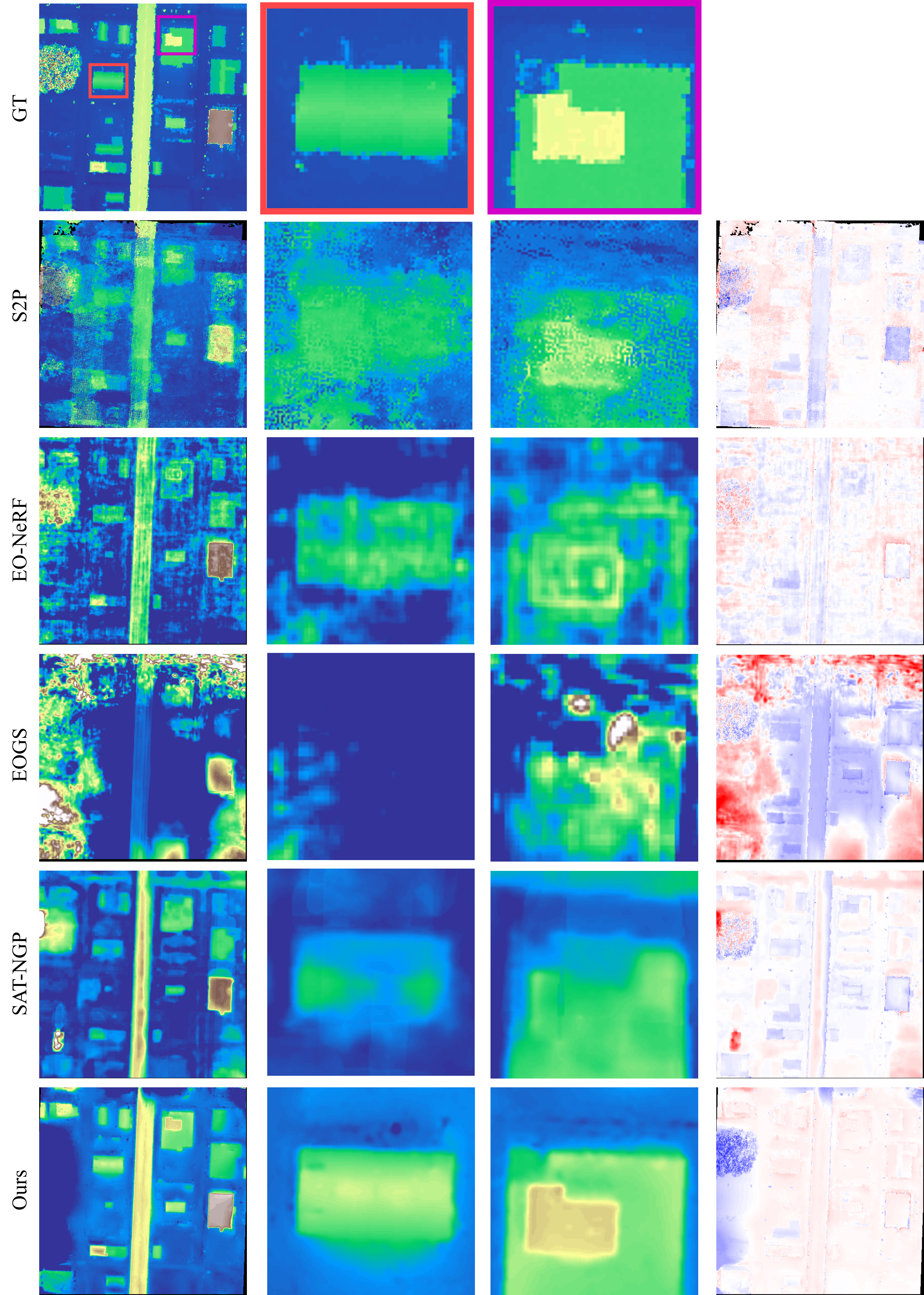}
    \caption{Geometric reconstruction visualization on the OMA 315 dataset. The fourth column shows the error map between each method's reconstructed DSM and the ground truth (GT), where red indicates overestimation and blue indicates underestimation of height values.}
    \label{fig:11}
\end{figure}
\newpage

\begin{figure}[ht]
    \centering
 \includegraphics[width=\linewidth]{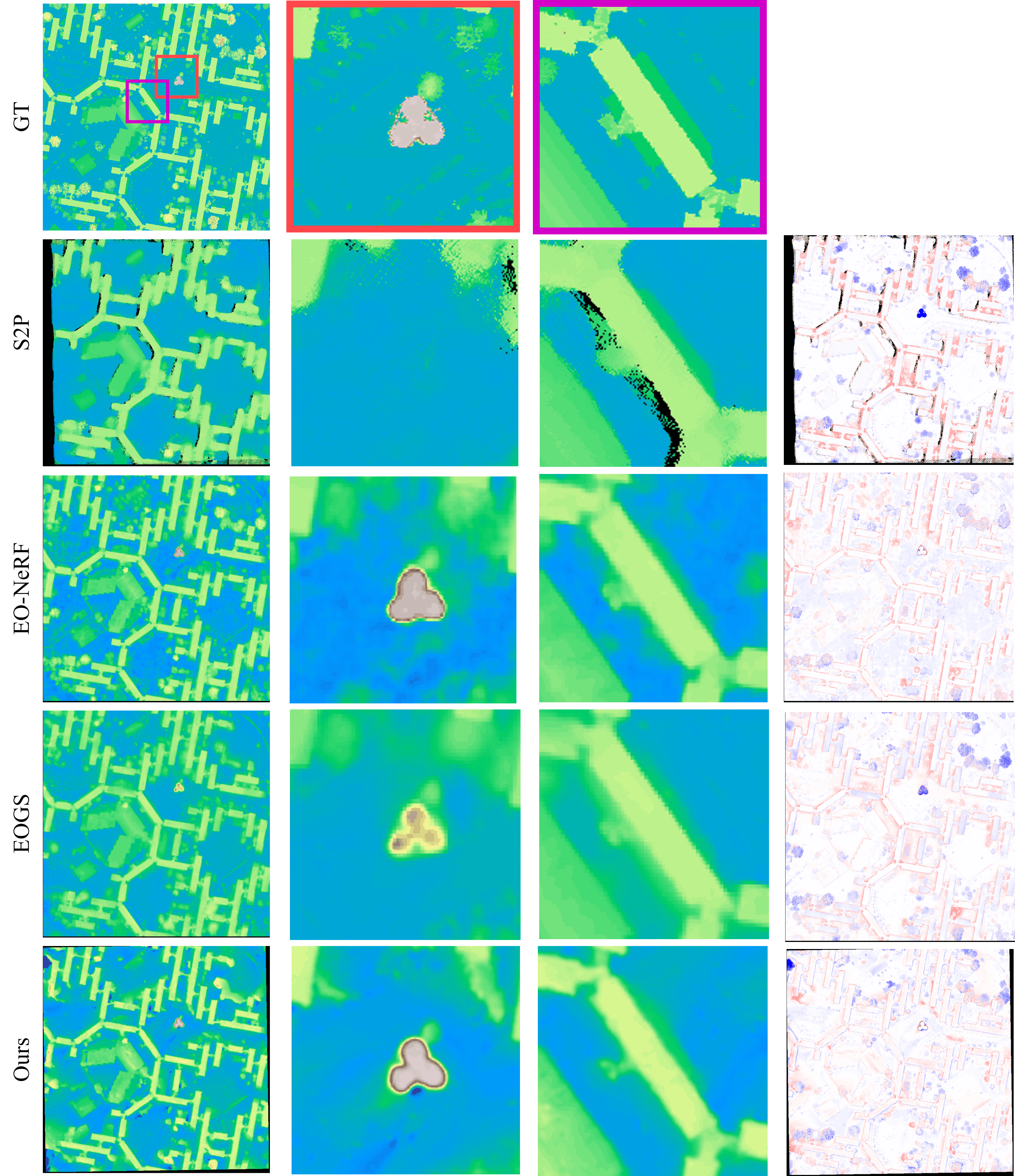}
    \caption{Geometric reconstruction visualization on the IARPA 001 dataset. The fourth column shows the error map between each method's reconstructed DSM and the ground truth (GT), where red indicates overestimation and blue indicates underestimation of height values.}
    \label{fig:12}
\end{figure}
\newpage

\begin{figure}[ht]
    \centering
 \includegraphics[width=\linewidth]{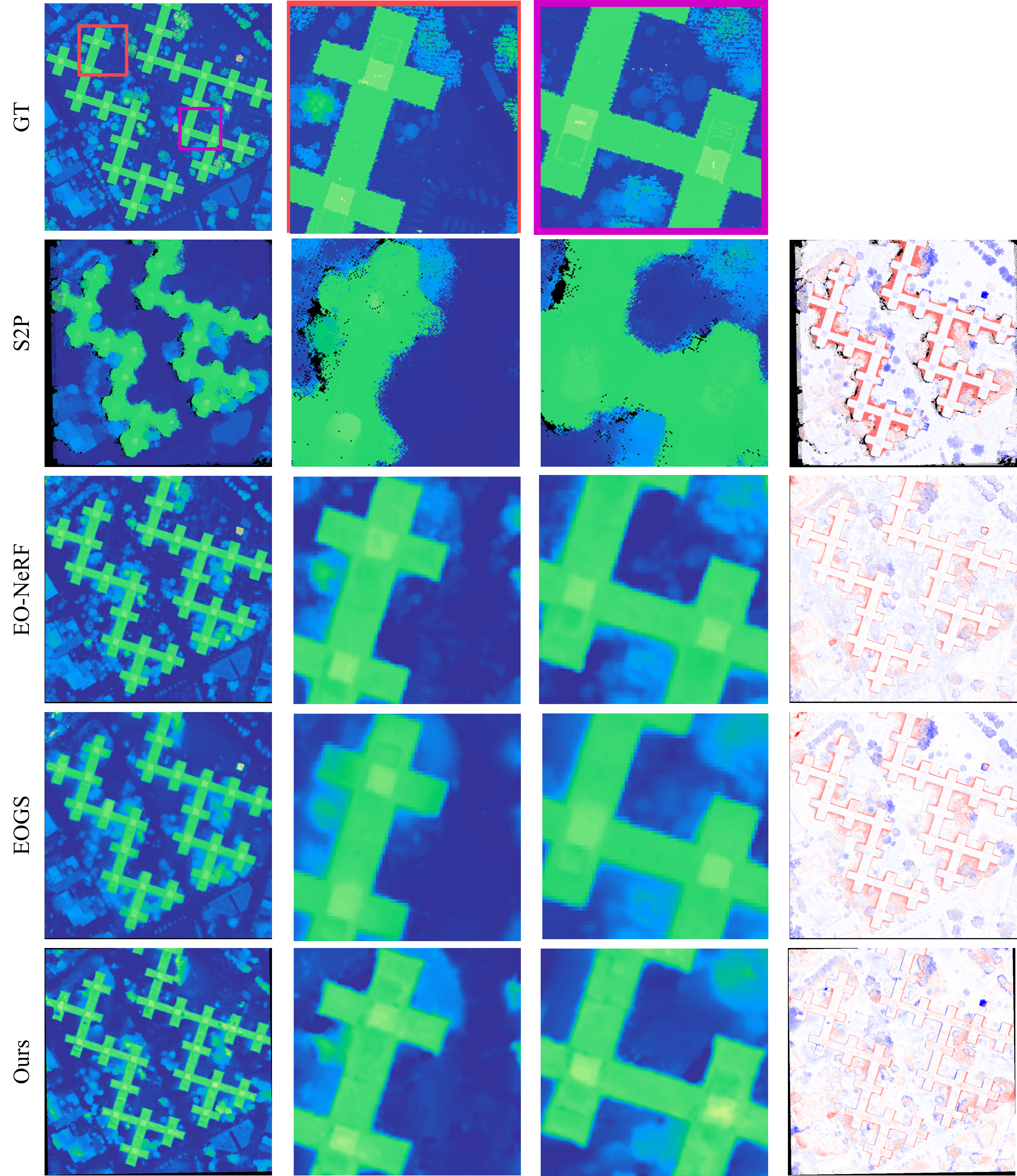}
    \caption{Geometric reconstruction visualization on the IARPA 002 dataset. The fourth column shows the error map between each method's reconstructed DSM and the ground truth (GT), where red indicates overestimation and blue indicates underestimation of height values.}
    \label{fig:13}
\end{figure}
\newpage

\begin{figure}[ht]
    \centering
 \includegraphics[width=\linewidth]{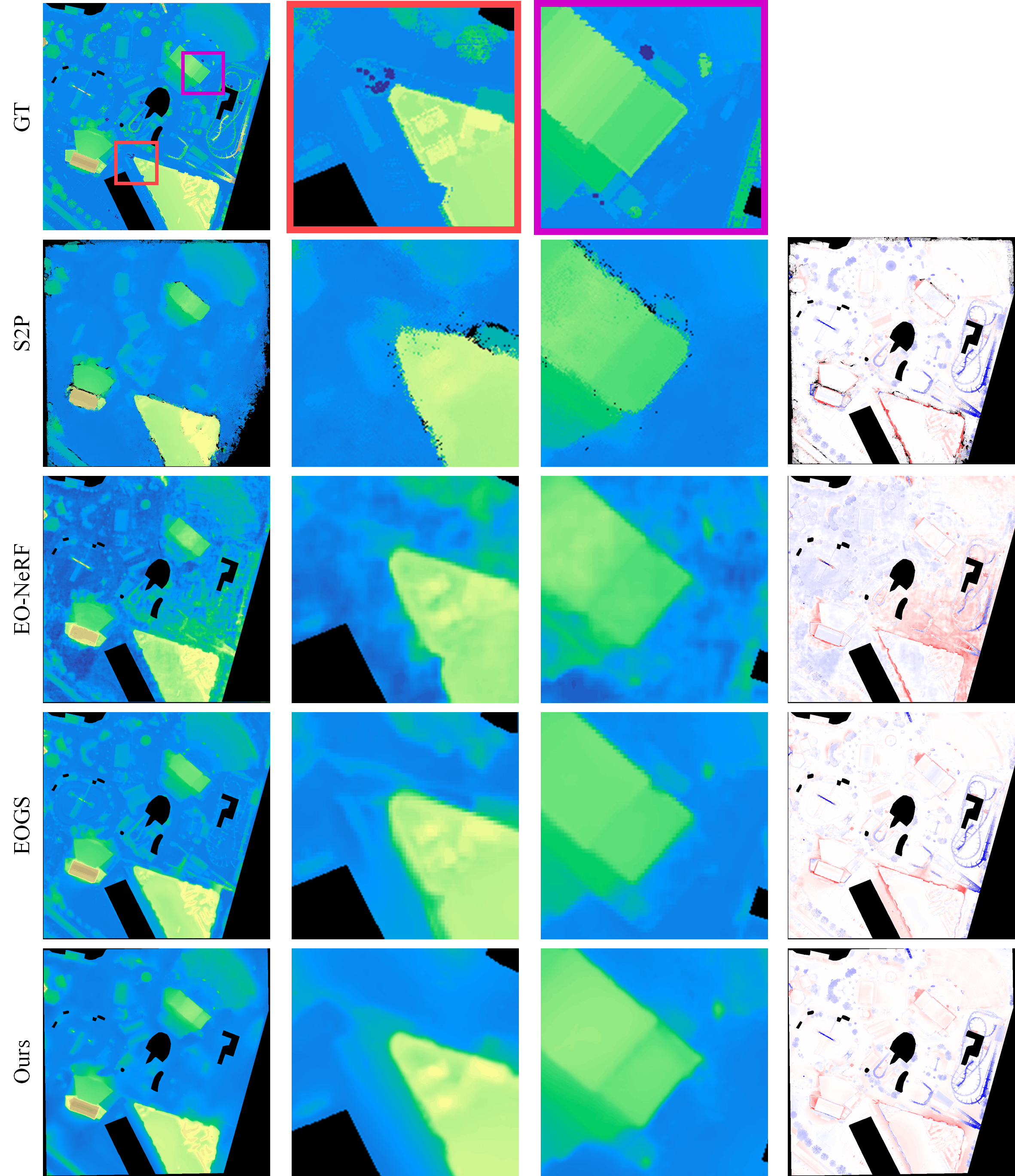}
    \caption{Geometric reconstruction visualization on the IARPA 003 dataset. The fourth column shows the error map between each method's reconstructed DSM and the ground truth (GT), where red indicates overestimation and blue indicates underestimation of height values.}
    \label{fig:14}
\end{figure}

\subsection*{A.2. Novel View Synthesis Results}
We provide comparative results of novel view synthesis in Figures 15-17.

\begin{figure}[ht]
    \centering
 \includegraphics[width=\linewidth]{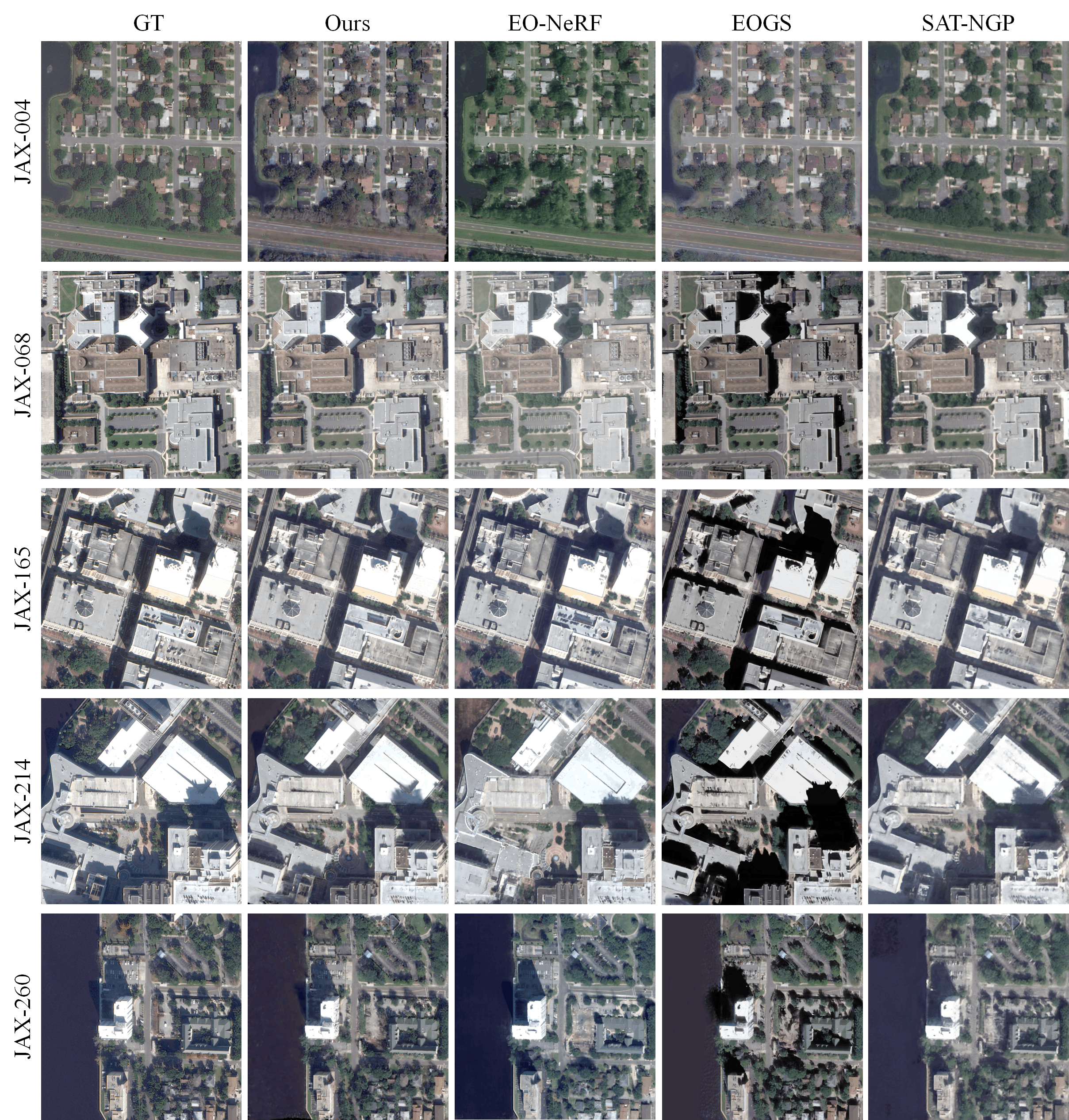}
    \caption{Visualization of novel view synthesis on 5 AOIs in the JAX region.}
    \label{fig:15}
\end{figure}
\newpage

\begin{figure}[ht]
    \centering
 \includegraphics[width=\linewidth]{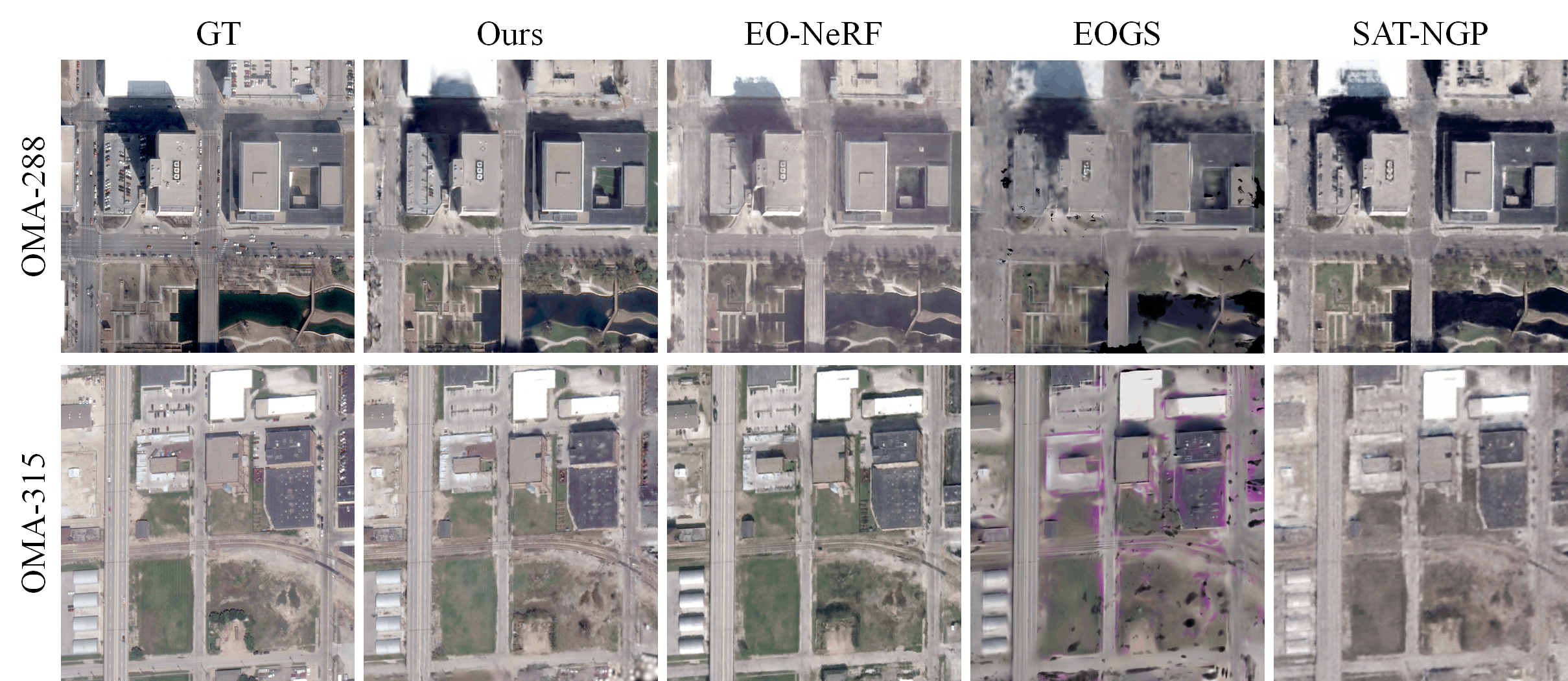}
    \caption{Visualization of novel view synthesis on 2 AOIs in the OMA region.}
    \label{fig:16}
\end{figure}
\newpage

\begin{figure}[ht]
    \centering
 \includegraphics[width=\linewidth]{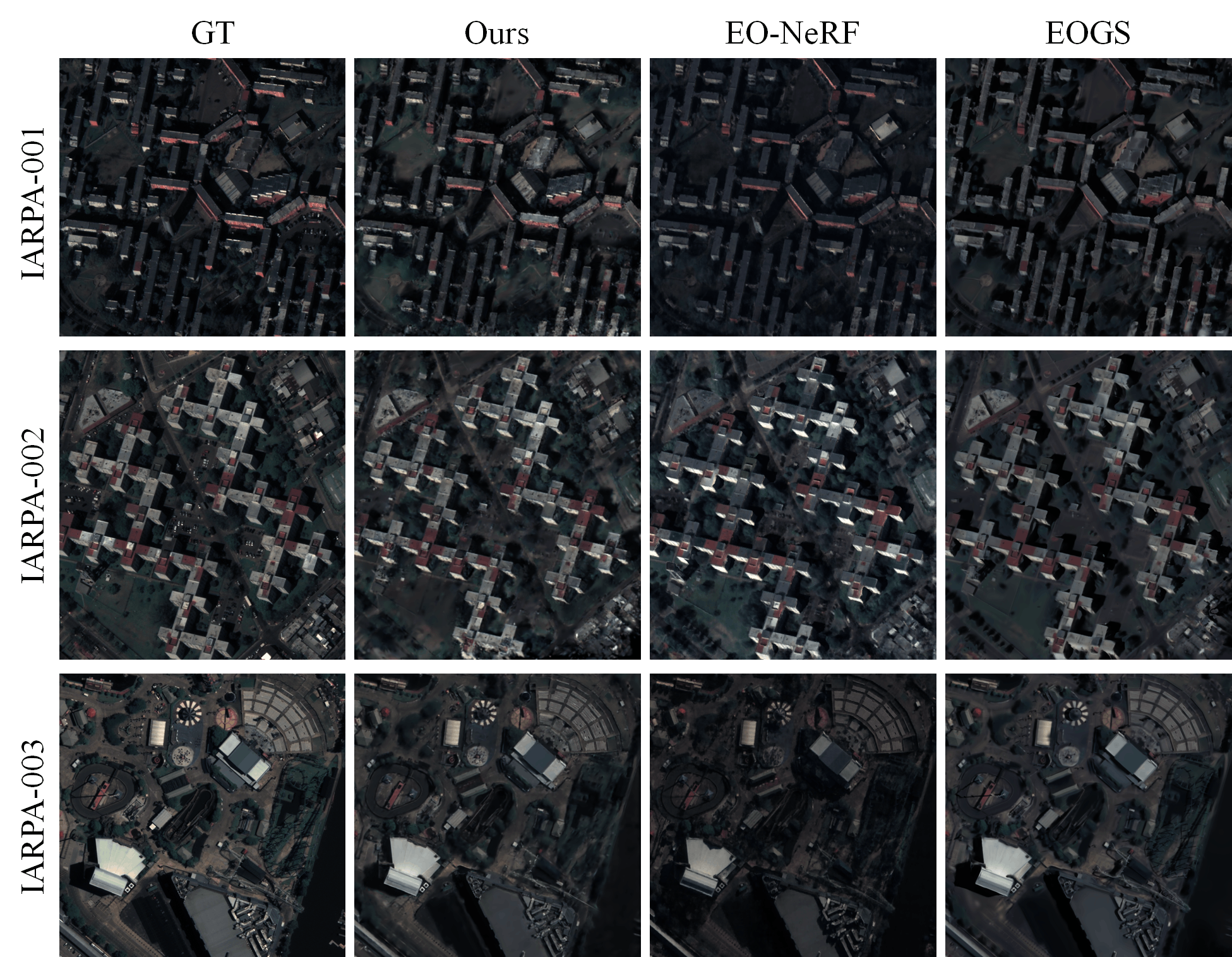}
    \caption{Visualization of novel view synthesis on 3 AOIs in the IARPA region.}
    \label{fig:17}
\end{figure}

\subsection*{A.3. Shadow and Albedo Decomposition Results}
We provide visualization results of shadow and albedo decomposition in Figures 18-22.

\begin{figure}[ht]
    \centering
 \includegraphics[width=\linewidth]{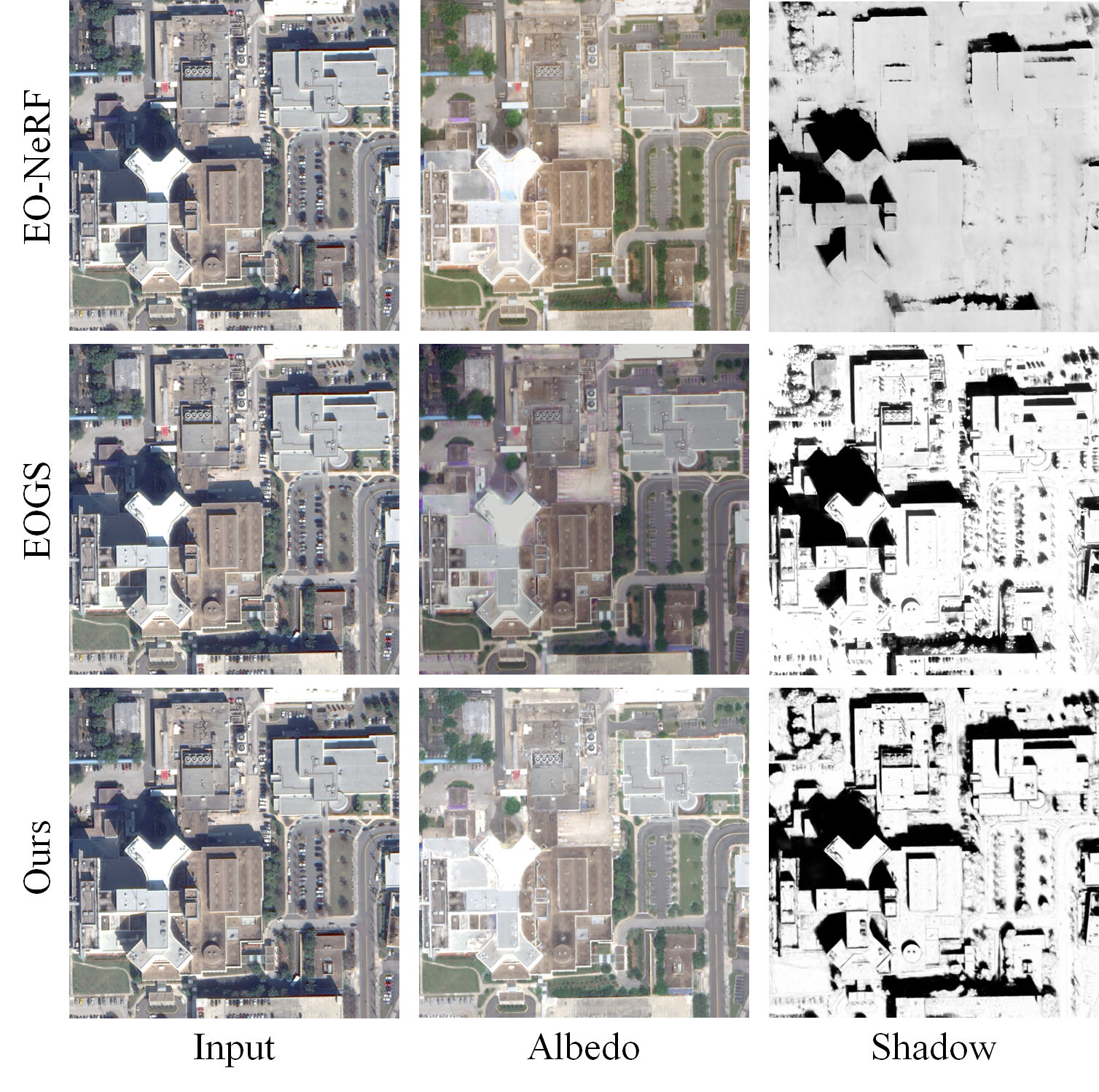}
    \caption{Shadow decomposition and albedo visualization on the JAX 068 dataset.}
    \label{fig:18}
\end{figure}
\newpage

\begin{figure}[ht]
    \centering
 \includegraphics[width=\linewidth]{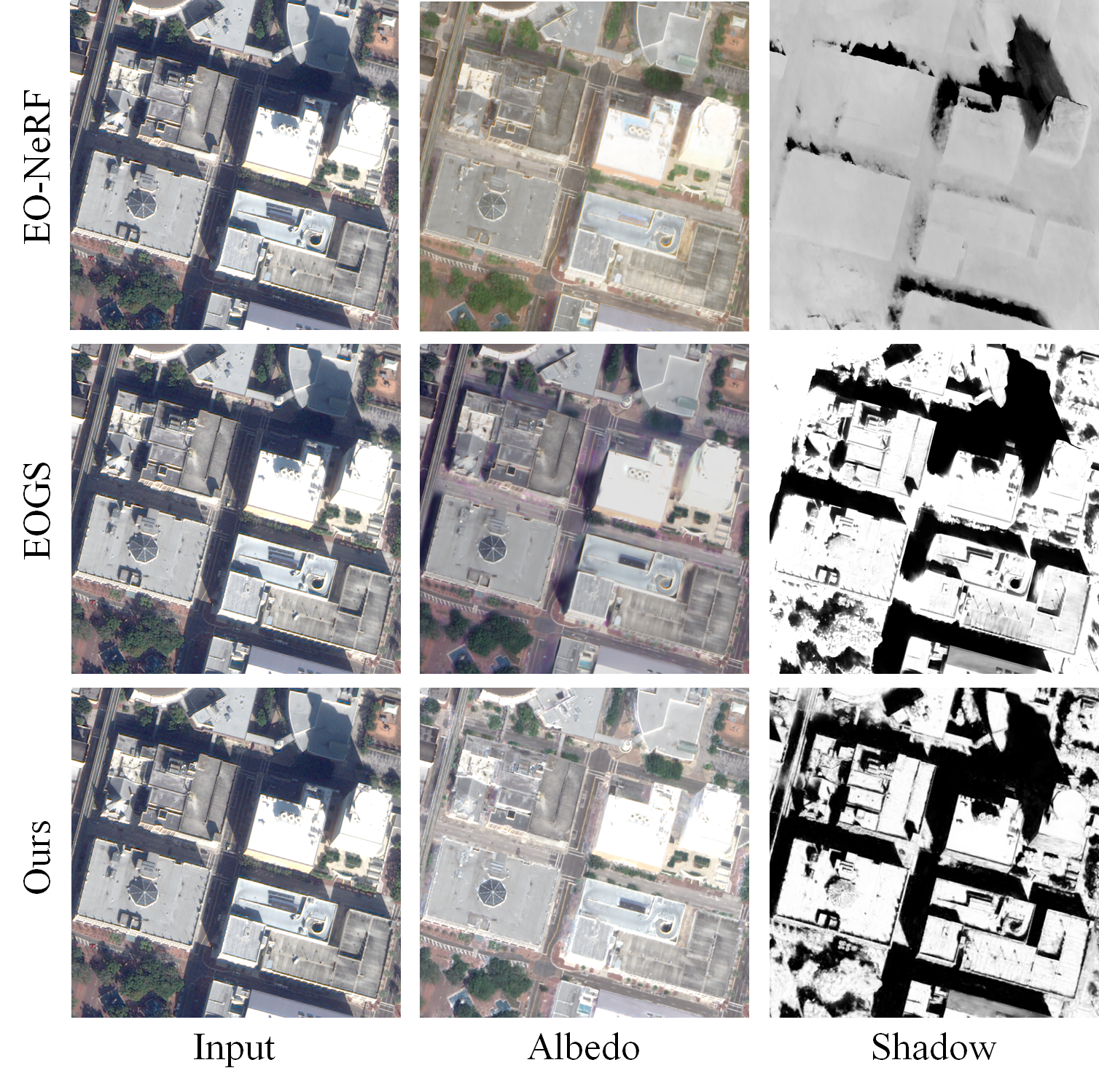}
    \caption{Shadow decomposition and albedo visualization on the JAX 165 dataset.}
    \label{fig:19}
\end{figure}

\begin{figure}[ht]
    \centering
 \includegraphics[width=\linewidth]{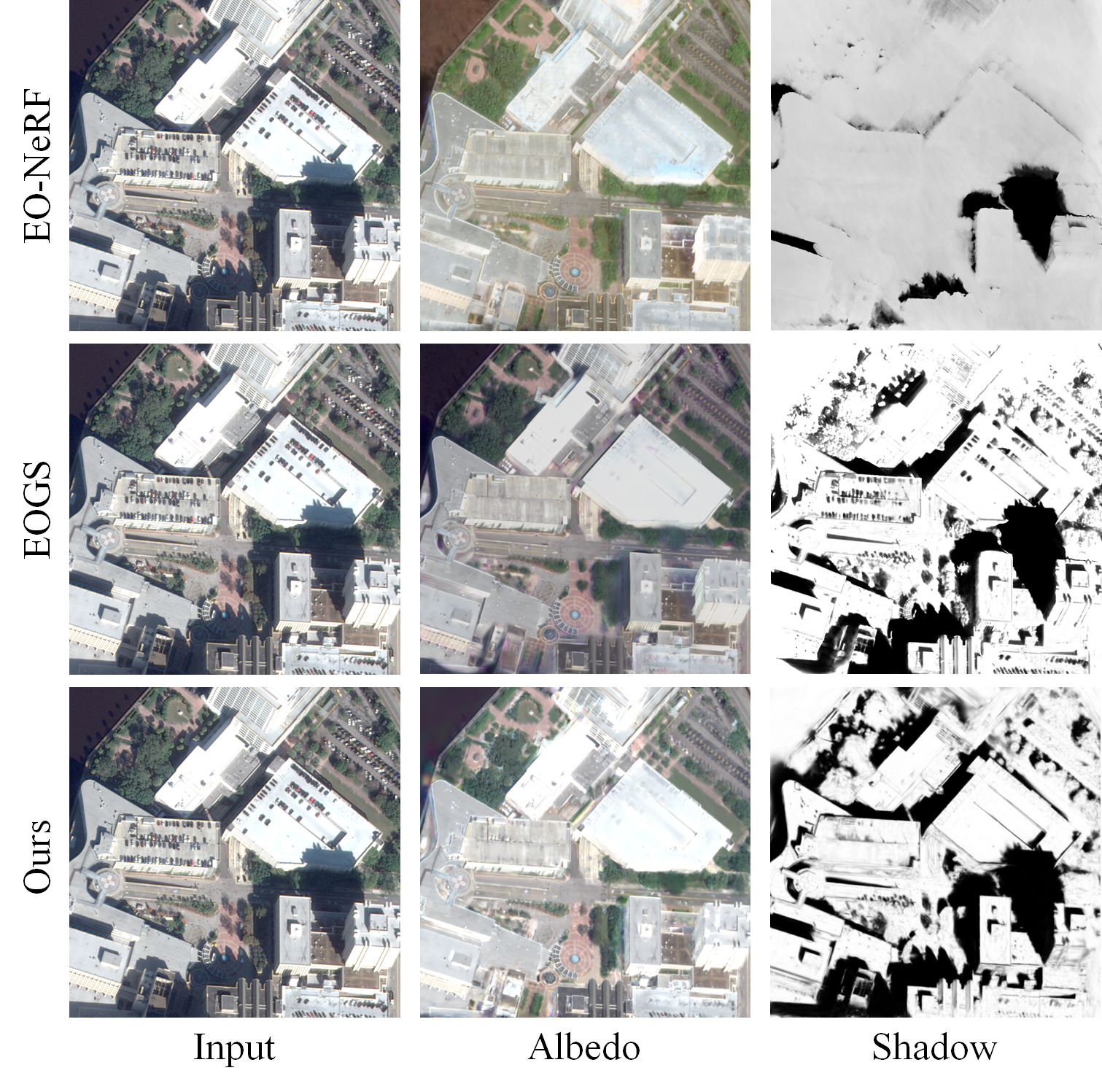}
    \caption{Shadow decomposition and albedo visualization on the JAX 214 dataset.}
    \label{fig:20}
\end{figure}
\newpage 

\begin{figure}[ht]
    \centering
 \includegraphics[width=\linewidth]{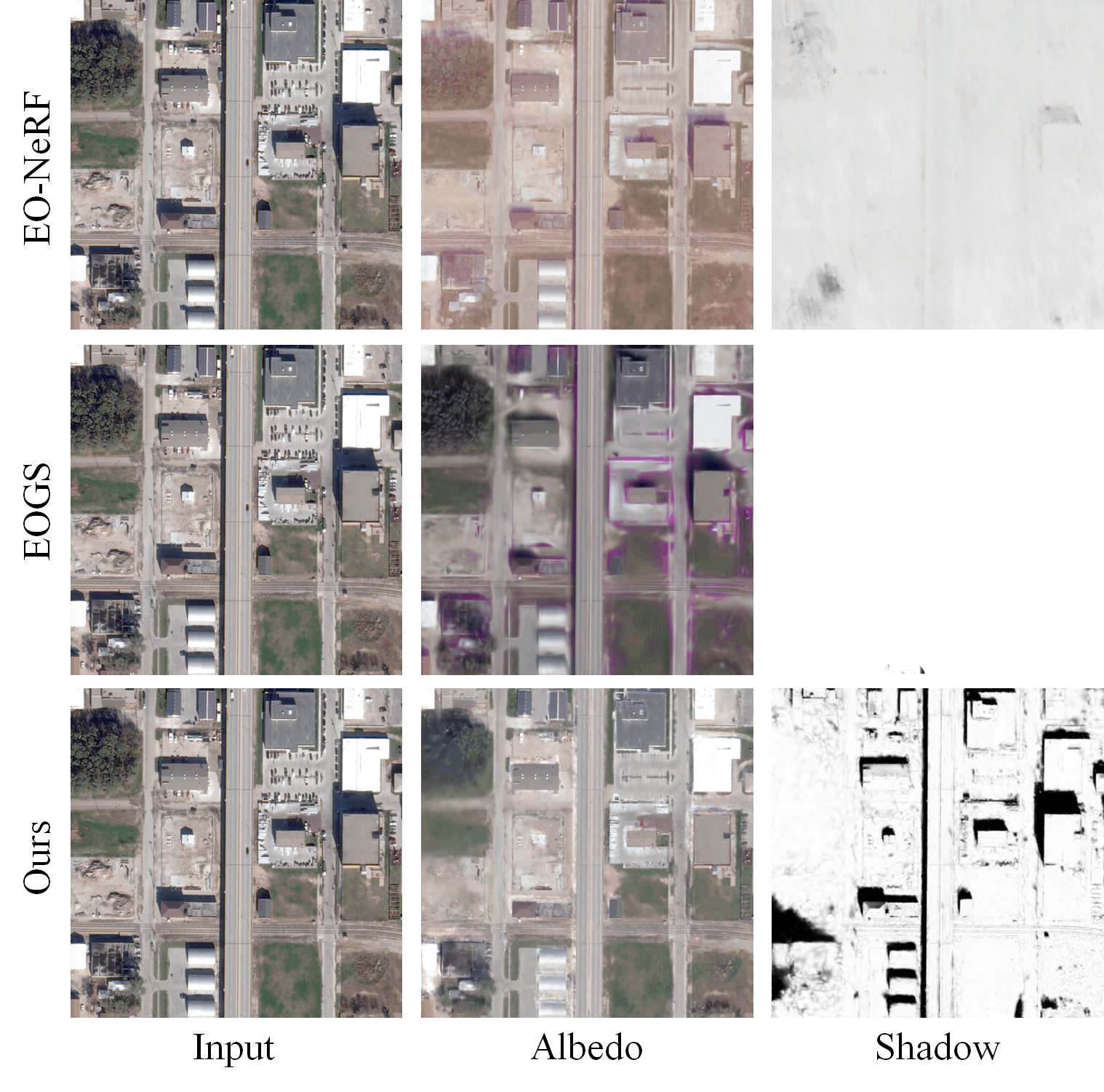}
    \caption{Shadow decomposition and albedo visualization on the OMA 315 dataset.}
    \label{fig:21}
\end{figure} 

\begin{figure}[ht]
    \centering
 \includegraphics[width=\linewidth]{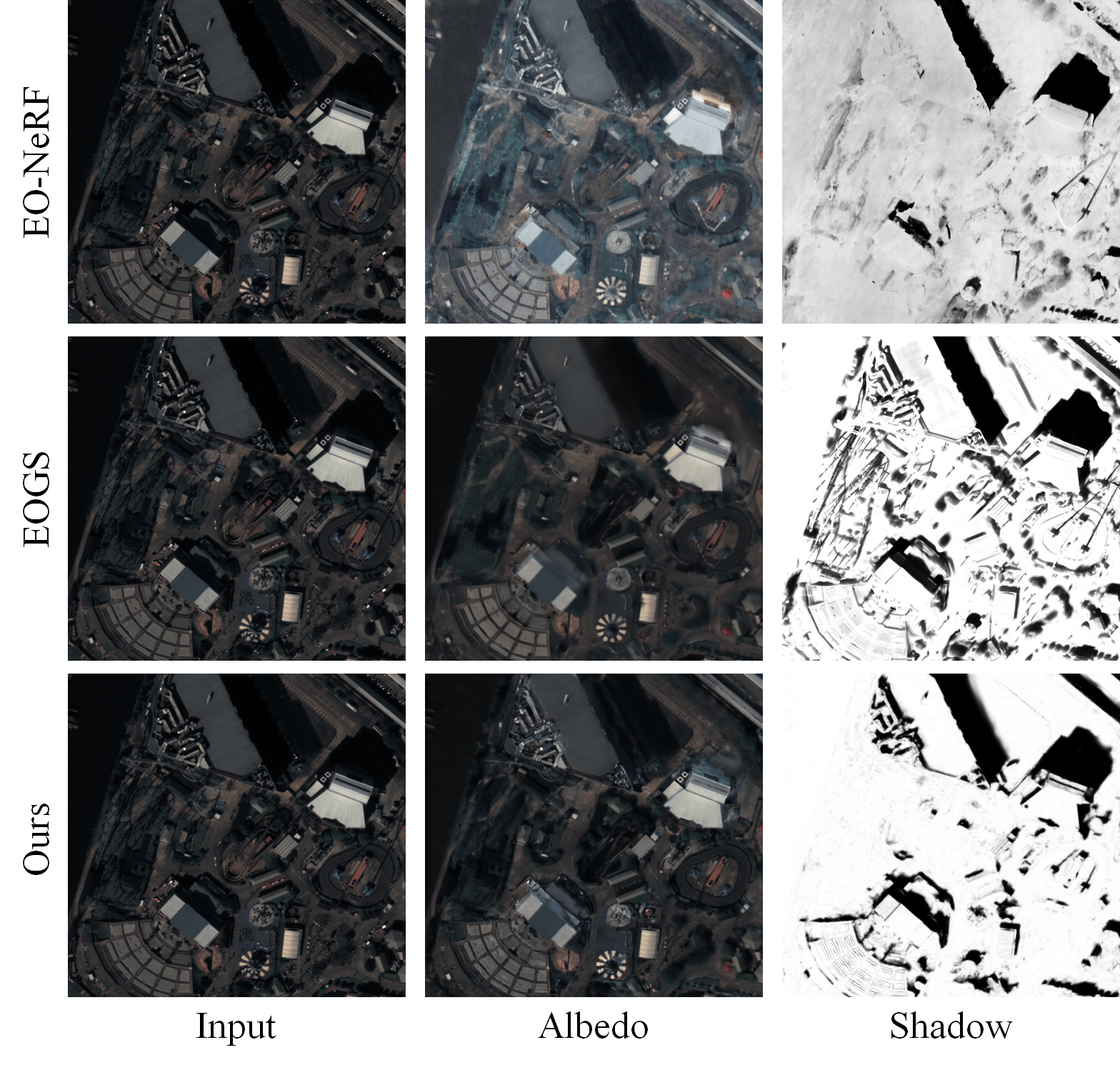}
    \caption{Shadow decomposition and albedo visualization on the IARPA 003 dataset.}
    \label{fig:22}
\end{figure}

\section*{B. Other Experiments}
We  evaluated the effect of incorporating FDRNet shadow map priors on reconstruction accuracy across different numbers of input views. As shown in \cref{tab:5}, the priors consistently improve performance when using limited views ($\leq$7). However, their effectiveness diminishes with more input views, eventually leading to performance degradation. This occurs because the inevitable false detections in shadow priors and inconsistencies among multiple priors introduce noise that adversely affects the reconstruction.

\begin{table}[ht]
\centering
\resizebox{\linewidth}{!}{\begin{tabular}{lccccc}
\toprule
MAE (m) & 3 views & 5 views & 7 views & 9 views & All views \\
\midrule
w/o Shadow mask & 3.30 & 2.40 & 1.96 & 1.64 & 1.41 \\
Shadow mask     & 2.67 & 2.28 & 1.89 & 1.70 & 1.47 \\
\bottomrule
\end{tabular}}
\caption{Impact of FDRNet shadow map priors on mean height MAE(m) across 5 AOIs in the JAX region under varying numbers of input views.
}
\label{tab:5}
\end{table}

\end{document}